\documentclass[10pt,twocolumn,letterpaper]{article}

\usepackage{cvpr}            
\usepackage{multirow}
\usepackage{multicol}
\usepackage{hyperref}
\usepackage{url}
\usepackage{orcidlink}
\usepackage{graphicx}
\usepackage{amsmath}
\usepackage{amssymb}
\usepackage{booktabs}
\usepackage{colortbl}
\usepackage{xcolor}
\usepackage{amsmath}
\usepackage{amsthm}
\usepackage{algorithm}
\usepackage{algpseudocode}
\usepackage{subcaption}
\usepackage{color, colortbl}
\usepackage{pifont}
\definecolor{LightCyan}{rgb}{0.88,1,1}
\definecolor{LightRed}{rgb}{1,0.88,0.88}
\definecolor{Gray}{rgb}{0.8,0.8,0.8}
\definecolor{LightGray}{rgb}{0.92,0.92,0.92}
\definecolor{LightPurple}{RGB}{226, 225, 254}
\definecolor{myblue}{RGB}{53,144,243}   % blue from screenshot
\definecolor{myred}{RGB}{232,118,89}    % red/orange from screenshot
\usepackage{xcolor}
\newtheorem{definition}{Definition}
\newtheorem{remark}{Remark}

\usepackage{enumitem}
\setlist{nosep, leftmargin=14pt}

\usepackage{mwe}
\usepackage{wrapfig}

\newcommand{\cmark}{\ding{51}}%
\newcommand{\xmark}{\ding{55}}%

\def\paperID{35018}
\def\confName{CVPR}
\def\confYear{2026}

\title{ManifoldGD: Training-Free Hierarchical Manifold Guidance for Diffusion-Based Dataset Distillation}

\author{Ayush Roy\\
University at Buffalo, SUNY
\and
Wei-Yang Alex Lee\\
University at Buffalo, SUNY
\and
Rudrasis Chakraborty\\
Lawrence Livermore National Lab
\and
Vishnu Suresh Lokhande\\
University at Buffalo, SUNY
}

\begin{document}
\maketitle
\begin{abstract}
In recent times, large datasets hinder efficient model training while also containing redundant concepts. Dataset distillation aims to synthesize compact datasets that preserve the knowledge of large-scale training sets while drastically reducing storage and computation. Recent advances in diffusion models have enabled training-free distillation by leveraging pre-trained generative priors; however, existing guidance strategies remain limited. Current score-based methods either perform unguided denoising or rely on simple mode-based guidance toward instance prototype centroids (IPC centroids), which often are rudimentary and suboptimal. We propose Manifold-Guided Distillation (ManifoldGD), a training-free diffusion-based framework that integrates manifold consistent guidance at every denoising timestep. Our method employs IPCs computed via a hierarchical, divisive clustering of VAE latent features, yielding a multi-scale coreset of IPCs that captures both coarse semantic modes and fine intra-class variability. Using a local neighborhood of the extracted IPC centroids, we create the latent manifold for each diffusion denoising timestep. At each denoising step, we project the mode-alignment vector onto the local tangent space of the estimated latent manifold, thus constraining the generation trajectory to remain manifold-faithful while preserving semantic consistency. This formulation improves representativeness, diversity, and image fidelity without requiring any model retraining. Empirical results demonstrate consistent gains over existing training-free and training-based baselines in terms of FID, $\ell_2$ distance among real and synthetic dataset embeddings, and classification accuracy, establishing ManifoldGD as the first geometry-aware training-free data distillation framework. The code is available on \url{https://github.com/AyushRoy2001/ManifoldGD}.
\end{abstract}    
\section{Introduction}
\label{sec:intro}

\begin{figure}[!t]
    \centering
    \includegraphics[width=\linewidth, keepaspectratio]{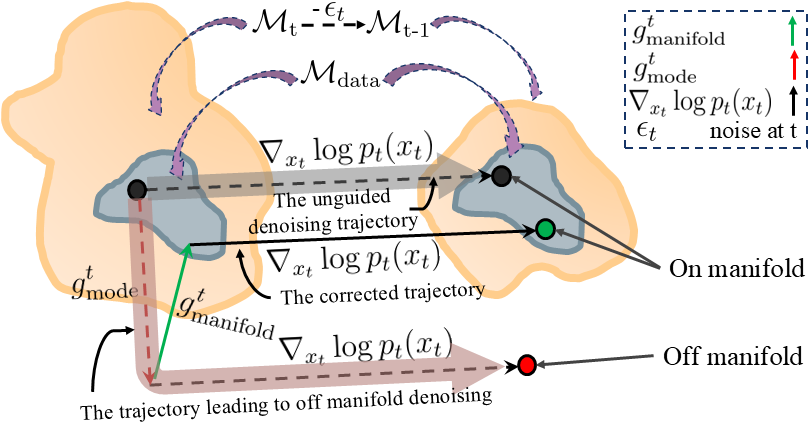}
    \vspace{-2em}
    \caption{\footnotesize\textbf{Manifold Guidance.} Overall denoising trajectory correction by manifold guidance ($g^t_\mathrm{manifold}$) to recitfy the off manifold component of mode guidance ($g^t_\mathrm{mode}$). $\mathcal{M}_\mathrm{data}$ is mebedded inside $\mathcal{M}_{t}$ and removing noise $\epsilon_t$ from $\mathcal{M}_{t}$ transforms it to $\mathcal{M}_{t-1}$. As $t \to 0$, $\mathcal{M}_{t} \to \mathcal{M}_\mathrm{data}$.} 
    \vspace{-2em}
    \label{fig:intro}
\end{figure}
The rapid progress of machine learning has been driven by increasingly large datasets and models, enabling state-of-the-art performance but posing severe challenges for researchers with limited computational or storage resources \cite{lokhande2022equivariance}. While model pruning \cite{ding2019centripetal,he2019filter,sharma2022rapid} and quantization \cite{chauhan2023post,chen2021towards} improve model efficiency, scalability from the data perspective is primarily addressed through \emph{data distillation} \cite{wang2018dataset,liu2022dataset} and \emph{coreset selection} \cite{welling2009herding,chen2012super,rebuffi2017icarl}. In modern computer vision, the explosive growth of labeled datasets \cite{slobogean2015bigger} makes training large models on millions of images computationally prohibitive \cite{lokhande2022equivariance, chytas2024pooling}. Dataset distillation tackles this by ``compressing'' the full dataset $\mathcal{D}$ into a much smaller synthetic subset $\mathcal{S}$ such that training on $\mathcal{S}$ yields performance comparable to $\mathcal{D}$. Earlier approaches, including coreset selection \cite{welling2009herding,chen2012super,rebuffi2017icarl} and gradient-matching methods \cite{cazenavette2022dataset,zhao2023dataset,wang2025emphasizing,yang2024neural,lee2024selmatch,mehta2023efficient}, optimize $\mathcal{S}$ so that the training trajectory or feature statistics of a model trained on $\mathcal{S}$ approximate those from $\mathcal{D}$. Despite their promise, these methods often rely on costly bi-level optimization, exhibit sensitivity across architectures, and struggle to capture rare modes of the data distribution \cite{chan2025mgd3,mehta2023efficient}.

The advent of large pre-trained generative models \cite{wang2024dim,zhang2023dataset} has transformed distillation: generative priors allow synthetic sampling and representational coverage of complex image manifolds. The generative model acts like a container storing the knowledge of $\mathcal{D}$ and using it to generate $\mathcal{S}$. By leveraging generative models, recent distillation works achieve higher fidelity, greater diversity and better downstream utility. Within this trend, diffusion-based distillation \cite{chan2025mgd3,gu2024minimaxdiff,su2024d4m,zou2025dataset,chen2025influence} has emerged as state-of-the-art. Yet, many generative approaches remain training-based: they require fine-tuning the generator \cite{su2024d4m,zou2025dataset}, optimizing synthetic images via bi-level loops \cite{gu2024minimaxdiff}, which raises cost and complexity. Moreover, none explicitly enforce that synthetic trajectories remain faithful to the latent data manifold, i.e., once guided toward data distribution modes, samples may drift off-manifold \cite{park2025temporal,he2023manifold}.

We explore a training free generative model based data distillation paradigm where achieve comparable performance training based methods while outperforming most of the existing methods. Our core contributions are as follows considering the aforementioned issues:
\begin{itemize}
  \item A fully \textbf{training-free} dataset distillation pipeline that uses only a pre-trained generator and is \textbf{inference-only}.
  \item A \textbf{layer‐wise hierarchical clustering} of VAE latent features to select IPC (images‐per‐class) centroids that represent the class modes in the VAE feature space. The IPC centroid selection from each level of the tree generated during hierarchical clustering covers coarse (levels closer to the root node) to fine modes (levels closer to the leaf nodes) without optimization.e generative model based data distillation.
  \item A \textbf{manifold‐guidance} strategy to offer trajectory correction introduced by mode‐guidance: guiding towards IPC centroids while constraining updates to the local latent tangent subspace to maintain data-manifold fidelity.
\end{itemize}

\section{Related Works}
\label{sec:related}

Dataset distillation has found importance for diverse applications such as continual learning \cite{zhao2023dataset,zhao2020dataset}, privacy preservation \cite{li2020soft,sucholutsky2021secdd}, neural architecture search \cite{zhao2020dataset,zhao2023dm}, and model interpretability \cite{loo2022efficient}. Early approaches relied on \emph{coreset selection} or \emph{gradient matching} to identify representative subsets \cite{DBLP:conf/icml/Welling09,sener2018coreset,pmlr-v162-kim22c,cazenavette2023glad}. While computationally efficient, these methods are often architecture-specific and limited in capturing complex, multimodal distributions.

\noindent
\textbf{Generative-model–based distillation.}  
The advent of pre-trained diffusion and latent generative models reshaped dataset distillation by enabling high-fidelity, conditional synthesis directly from generative priors. Training-based diffusion methods such as DM \cite{zhao2023dm}, Min–Max Diffusion \cite{gu2024minimaxdiff}, D$^{4}$M \cite{su2024d4m}, and GLAD \cite{cazenavette2023glad} fine-tune the generative model or optimize synthetic images via a min–max or gradient-matching objective to balance diversity and representativeness. Although effective, these approaches are computationally expensive and often require additional optimization stages for each dataset. 

\noindent
\textbf{Training-free diffusion methods.}  
More recent works have shifted toward \emph{training-free} paradigms that avoid fine-tuning. Diffusion Transformers (DiT) \cite{peebles2023dit} and Latent Diffusion Models (LDM) \cite{rombach2022ldm} can generate synthetic datasets directly, but unguided sampling often yields semantically diffuse or redundant samples. Mode-Guided Diffusion (MGD) \cite{chan2025mgd3} introduces cluster-based guidance to steer sampling toward class centroids, improving sample diversity and semantic consistency. However, MGD and similar methods rely purely on Euclidean attraction, leading to potential off-manifold drift and degraded fidelity.

A few concurrent efforts address more principled guidance mechanisms. Information-Guided Diffusion \cite{ye2025information} uses an information-theoretic objective that depends on a separately trained classifier; Influence-Guided Diffusion \cite{chen2025influence} computes influence scores from pre-trained classifiers to steer the denoising trajectory; and text-guided methods \cite{zou2025dataset} or OOD-aware generators \cite{gao2025good} rely on additional conditioning or discriminators. While these methods enhance control, they \textbf{require auxiliary networks} beyond the diffusion backbone. In contrast, our proposed ManifoldGD is a fully training-free method that uses only a pre-trained diffusion model and its accompanying VAE feature space. By constraining the guidance direction to the tangent space of a locally estimated diffusion manifold, ManifoldGD enforces geometric consistency, preventing off-manifold drift and improving representativeness, diversity, and without any additional supervision or model fine-tuning.
\section{Method}
\label{sec:method}

\begin{figure*}[!t]
    \centering
    \includegraphics[width=0.9\linewidth, keepaspectratio]{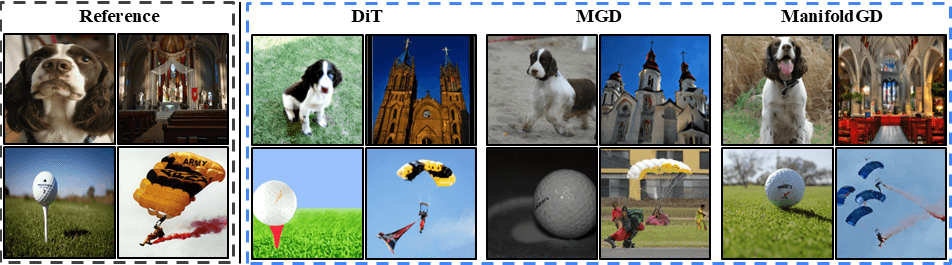}
    \caption{\footnotesize\textbf{Qualitative samples generated by DiT \cite{peebles2023dit}, MGD \cite{chan2025mgd3}, and ManifoldGD.} The samples generated by ManifoldGD have better image structure and quality (eg. dog image of MGD is having legs in unusual position, dog image generated by DiT is blurry. Similarly, the buildings have uncommon structure for MGD. The ball image generated by DiT is of poor quality).} 
    \label{fig:qualitative}
    \vspace{-1.5em}
\end{figure*}

\subsection{Preliminaries}
\label{sec:problem_statement}

\noindent{\bf Dataset Distillation.} Let the real dataset be $\mathcal{D} = \{(x_i, y_i)\}_{i=1}^{N}, \quad x_i \in \mathbb{R}^D,\; y_i \in \{1, \dots, C\}$,
where $x_i$ denotes the input sample and $y_i$ the associated class label.  
Under the standard Empirical Risk Minimization (ERM) \cite{shalev2014understanding}, a classifier $f_\theta$ with parameters $\theta$ is trained by minimizing the expected loss $\mathcal{L}_{\mathcal D}(\theta) = \frac{1}{N}\sum_{i=1}^N \ell\big(f_\theta(x_i), y_i\big)$,
where $\ell(\cdot,\cdot)$ denotes the task-specific loss function.  

Dataset distillation seeks to construct a significantly smaller synthetic dataset $\mathcal{S} = \{(\tilde{x}_j, \tilde{y}_j)\}_{j=1}^{M}$, $M \ll N$,
such that training on $\mathcal{S}$ yields parameters $\theta^{*}_{\mathcal{S}} = \arg\min_{\theta}\frac{1}{M}\sum_{j=1}^{M}\ell\big(f_\theta(\tilde{x}_j), \tilde{y}_j\big)$, whose downstream performance approximates that of $\theta^{*}_{\mathcal{D}}$. Formally, synthetic dataset minimizes the discrepancy in population risk 
\begingroup
\footnotesize
\begin{align}
\mathcal{S}^{*}
=\arg\min_{\mathcal{S}}
\left|
\operatorname*{\mathbb{E}}_{(x,y)\sim\mathcal{D}}
\!\ell\!\left(f_{\theta_{\mathcal{S}}^{*}}(x),y\right)-\operatorname*{\mathbb{E}}_{(x,y)\sim\mathcal{D}}
\!\ell\!\left(f_{\theta_{\mathcal{D}}^{*}}(x),y\right)
\right|    
\end{align}
\endgroup
In this way, the distilled dataset is explicitly chosen to preserve the predictive behavior of a model trained on the full dataset.

\noindent{\bf Decomposing the Score in Conditional Diffusion.}
\label{sec:diffusion_background}
A diffusion model defines a forward noising process $\{x_t\}_{t=0}^{T}$ that gradually perturbs data samples $x_0 \!\sim\! p_{\mathrm{data}}(x)$ into Gaussian noise $x_T \!\sim\! \mathcal{N}(0, I)$. The corresponding reverse (denoising) process is parameterized by a time-dependent Stein's score function $s_\theta(x_t, t)$ \cite{chan2024tutorial} that is trained to match the log-probability density of $x_t$, that is $\nabla_{x_t} \log p_t(x_t)$. 
The score $s_\theta(x_t, t)$ defines a vector field on the sample space that points towards regions of higher probability under the marginal distribution $p_t(x_t)$ at noise level $t$. At each denoising step, updating $x_t$ in the direction of $s_\theta(x_t, t)$ incrementally increases its likelihood, projecting the sample onto the evolving data manifold and thereby performing progressive denoising. 

For conditional generation, we consider the conditional density $p_t(x_t|c)$, where $c$ denotes an external conditioning signal. In the context of data distillation, $c$ corresponds to an IPC centroid, that is, a representative $\mathrm{mode}$ associated with a class. Such centroids (or $\mathrm{mode}$s) may be obtained through a variety of procedures, one example being the extraction of class-representative modes in a variational autoencoder (VAE) feature space (see \cref{sec:practical_implementation} for a detailed discussion). Under this formulation, the score of the conditional distribution admits the standard decomposition:
{\small
\begin{align}
\label{eq:score_decomp}
\nabla_{x_t} \log p_t(x_t \mid c)
&=
\underbrace{
    \nabla_{x_t}\log p_t(x_t)
}_{\textcolor{myblue}{\text{(1) Marginal Denoising}}}+
\underbrace{
    \nabla_{x_t}\log p_t(c \mid x_t)
}_{\textcolor{myred}{\text{(2) Mode Guidance}}}
\end{align}}
Here, the term 
\textcolor{myblue}{\text{(1) Marginal Denoising}} 
drives the dynamics toward regions of high data density. In the limit $t \to 0$, as the injected noise vanishes, the transition density $p_t$ converges to the underlying data distribution $p_{\mathrm{data}}$~\cite{ho2020denoising}. The term 
\textcolor{myred}{\text{(2) Mode Guidance}} 
introduces a semantic correction by enforcing attraction toward the conditional mode associated with the signal $c$. Weighted, classifier-based, and prototype-based guidance can be interpreted as variants of this mechanism~\cite{chung2024cfg++,sadat2024no}.  Conceptually, 
\textcolor{myblue}{\text{(1) Marginal Denoising}} 
recovers the coarse geometric structure induced by the diffusion prior, whereas 
\textcolor{myred}{\text{(2) Mode Guidance}} 
steers the sampling trajectory toward the conditional distribution, that is, the semantic mode of the corresponding class. When this attraction is excessively strong or misaligned with the intrinsic manifold geometry, the resulting trajectory may deviate from the true data manifold~\cite{park2025temporal,he2023manifold} (see ~\cref{fig:intro}).

\subsection{Manifold-Aware Mode Guidance in Conditional Diffusion}
\label{sec:mode_guidance}
\noindent {\bf Manifold Considerations in Mode Guidance.} The mode guidance term in \cref{eq:score_decomp} can be expressed as a kernel affinity between between the diffusion sample $x_t$ and the modes $c$. That is,  $k_\phi(x_t, c) = \exp\!\big(-\phi(\|x_t - c\|_2)\big)$
where $\phi(\cdot)$ defines a positive semi-definite potential function that governs the structure of the affinity
\footnote{For example, $\phi(r)=\tfrac{r^2}{2\sigma_t^2}$ corresponds to the Gaussian (RBF) kernel,  
$\phi(r)=\tfrac{r}{\sigma_t}$ to the Laplace kernel, and  
$\phi(r)=\log(1 + \tfrac{r^2}{2\sigma_t^2})$ to the inverse multi-quadric (IMQ) kernel.}. The resulting log-posterior gradient is expressed as $\nabla_{x_t} \log p_t(c|x_t)=
-\phi'(\|x_t - c\|_2)\,
\frac{x_t - c}{\|x_t - c\|_2}$, yielding the generalized mode-guidance vector, 
$g_{\mathrm{mode}}^t
=
-\phi'(\|x_t - c\|_2)\,
\frac{x_t - c}{\|x_t - c\|_2}$.
Under the quadratic potential $\phi(r)=\tfrac{r^2}{2\sigma_t^2}$, the expression reduces to the standard Gaussian form $g_{\mathrm{mode}}^t = -\tfrac{1}{\sigma_t^2}(x_t - c)$, aligning with existing mode-guided diffusion formulations~\cite{chan2025mgd3,su2024d4m}. The complete denoising step can be expressed as,
\begin{align}
\label{eq:kernel_overall_denoising}
x_{t-1}
= x_t
+ \eta_t \big[ s_\theta(x_t, t) + g^t_{\mathrm{mode}} \big]
+ \sqrt{\beta_t}\,\epsilon_t
\end{align}
\vspace{-1.0em}

Mode guidance, using vectors like $g^t_{\mathrm{mode}} = -\tfrac{1}{\sigma_t^2}(x_t - c) $ assumes that this direction is meaningful in the ambient Euclidean space. The true generative manifold is typically a curved, lower-dimensional subset of the ambient space \cite{park2025temporal,he2023manifold}. Prior work shows that modeling group actions tied to nuisance variables enables representations that separate signal from nuisance structure on such manifolds \cite{lokhande2022equivariance}. Let $\mathcal{M} \subseteq \mathbb{R}^d$ denote the ambient data space. We define:
\begin{itemize}
\item The true data manifold $\mathcal{M}_{\mathrm{data}}$, a low-dimensional submanifold on which $p_{\mathrm{data}}(x)$ is supported.
\item The diffusion manifold $\mathcal{M}_t$ at timestep $t$, supporting the noisy data marginal $p_t(x_t)$. By construction, $\mathcal{M}_{\mathrm{data}} \subseteq \mathcal{M}_t$, and $\mathcal{M}_t \to \mathcal{M}_{\mathrm{data}}$ as $t \to 0$.
\end{itemize}
Then, at any  $x_t \in \mathcal{M}_t$ we have the tangent space $\mathcal{T}_{x_t}\mathcal{M}_t$ and  normal space  $\mathcal{N}_{x_t}$ defined using, \nobreak\vspace{-0.5em}
\begin{definition}[Tangent Score Function]
The tangent score at $x_t$ is $s_r(x_t) := \nabla_{x_t} \log p_{\mathcal{M}_t}(x_t) \ \in \mathcal{T}_{x_t}\mathcal{M}_t$.
\end{definition}
\begin{definition}[Normal Space]
The normal space at $x_t$ is $\mathcal{N}_{x_t} = \{ n_t \in \mathbb{R}^d : n_t \perp \mathcal{T}_{x_t}\mathcal{M}_t \}$.
\end{definition}
The off-manifold component is quantified by projection onto the normal space:
$$
\langle g_t^{\mathrm{mode}}, n_t \rangle = -\phi'(\|x_t - c\|_2)\, \bigg\langle \frac{x_t - c}{\|x_t - c\|_2},\, n_t \bigg\rangle, n_t \in \mathcal{N}_{x_t},
$$
where $\phi(\cdot)$ is the mode potential in affinity-based guidance. As diffusion proceeds ($t \to 0$), $p_t(x_t)$ becomes sharply concentrated near $\mathcal{M}_t \to \mathcal{M}_{\mathrm{data}}$, so even small normal components can substantially reduce likelihood under $p_{\mathrm{data}}(x)$.

\noindent {\bf Manifold-Constrained Mode Updates. } Unconstrained guidance $g_{\mathrm{mode}}^t$ enforces semantic attraction but violates manifold geometry, yielding invalid generations \cite{park2025temporal,he2023manifold}. Manifold-corrected guidance, typically via tangent space projection, maintains local fidelity to $\mathcal{M}_t$ and $\mathcal{M}_{\mathrm{data}}$. We decompose guidance into tangent and normal components via orthogonal projection onto the estimated diffusion manifold $\mathcal{M}_t$. Let $P_{\mathcal{T}_t}$ and $P_{\mathcal{N}_t} = I - P_{\mathcal{T}_t}$ \cite{absil2008optimization} denote orthogonal projectors onto $\mathcal{T}_{x_t}\mathcal{M}_t$ and its orthogonal complement, respectively. The tangential component $P_{\mathcal{T}_t} g^t_{\mathrm{mode}}$ aligns the update with the manifold while preserving local validity of the denoising process.
\vspace{-0.5em}
\begin{remark}
\label{remark_1}
While strict tangent projection ensures geometric consistency, it risks over smoothing by constraining exploration of valid manifold regions. This trade-off between geometric fidelity and sample diversity is particularly relevant given the curvature of real data manifolds and the limitations of local linear approximations.
\end{remark}
\vspace{-0.5em}
Alternatively, explicit normal component subtraction provides independent control over semantic attraction and manifold correction:
\begin{align}
    g^t_{\mathrm{manifold}}(x_t; c) = g^t_{\mathrm{mode}}(x_t; c) - P_{\mathcal{N}_t} g^t_{\mathrm{mode}}(x_t; c)
\end{align}
where the weighting parameters control the relative influence of semantic guidance and geometric correction, respectively. Thus, the full manifold-conditioned scoring function is given by $s^t_{\mathrm{manifold}}(x_t) = s_\theta(x_t, t) + g^t_{\mathrm{manifold}}$
and the corresponding sampling step at each reverse diffusion iteration is $x_{t-1} = x_t + \eta_t s^t_{\mathrm{manifold}}(x_t) + \sqrt{\beta_t} \epsilon_t$. This framework enables fine-grained control over manifold consistency and semantic alignment.

\subsection{Algorithmic Details for Manifold-Projected Correction}
\label{sec:practical_implementation}
In this section, we summarize the practical steps of our manifold-projected correction method for enforcing tangent-aligned guidance in diffusion-based dataset distillation. A full algorithmic specification is provided in the supplementary material.

\noindent {\bf \underline{Step (1)}}: \textit{VAE Latent Encoding and Diffusion.} Following prior works \cite{gu2024minimaxdiff, chan2025mgd3, lokhande2022equivariance}, each input image is encoded into a latent feature via a variational autoencoder (VAE), yielding $x_t$ that follows the diffusion process in latent space. A pretrained diffusion model (e.g., DiT) then performs the reverse denoising to map $x_t$ back toward the clean latent distribution approximating $p_{\mathrm{data}}$.

\noindent {\bf \underline{Step (2)}}: \textit{Building Mode-Guided Coresets (IPCs) and Local Manifolds.} We construct a coreset of IPC centroids $c_s$ via divisive (bisecting $k$-means) hierarchical clustering on class-wise VAE latents and use these centroids to build time-dependent local manifold $\mathcal{M}_t$  aligned with the diffusion noise level. Let the divisive tree leaves at depth $d$ be $\mathcal L_d$ for $d\!=\!0,\dots,L$ (root depth $0$). Given a start level $s_{\mathrm{start}}\in[0,L]$ (controls coarse→fine bias) and IPC budget $K$, we (i) pick one node per level in a coarse→fine sweep beginning at $s_{\mathrm{start}}$ and going down to the root, and (ii) if more IPCs are required, we fill the remaining quota by randomly selecting from the leaf nodes. This yields a deterministic coreset that favors global/generic modes when classes heavily overlap (higher $s_{\mathrm{start}}$) and progressively adds finer/specific modes. For each selected $c_s$, we define its neighborhood $\mathcal{N}_s$ (local region in latent space) which captures structure similar to $c_s$, so as to cover both broad and fine-grained variations of the data.

\noindent {\bf \underline{Step (3)}:} \textit{Constructing the Noisy Local Manifold at Diffusion Timestep $t$.} To match the current noise level in the diffusion process (at time $t$), we ``forward-diffuse" all points in $\mathcal{N}_s$ by adding Gaussian noise with variance $(1-\bar{\alpha}_t)$. The resulting set, $\mathcal{M}_t^{(s)}$, approximates the structure of the data manifold at this particular noise scale.
\begin{align}
\label{eq:Mt_construct}
\mathcal{M}_t^{(s)} = \mathcal{N}_s + \epsilon_t, 
\quad
\epsilon_t \sim \mathcal{N}\!\big(0,\,(1-\bar{\alpha}_t)I\big)
\end{align}
$\mathcal{M}_t^{(s)}$ represents the diffusion manifold at timestep $t$ that smoothly embeds $\mathcal{N}_s \subseteq \mathcal{M}_{\mathrm{data}}$ as $t\!\to\!0$.  

\noindent {\bf \underline{Step (4):}} \textit{Estimating Local Geometry (Tangent/Normal Subspaces) and Manifold Correction.}  Given a latent sample $x_t$, we identify its $K_t$ nearest neighbors within the local manifold patch $\mathcal{M}_t^{(s)}$ and compute the empirical covariance $C_t$ of these neighbors. The leading $d$ eigenvectors of $C_t$ define a basis for the tangent subspace $\mathcal{T}_{x_t}\mathcal{M}_t$, while the remaining orthogonal directions form the normal subspace $\mathcal{N}_{x_t}$. We construct the projection operators $P_{\mathcal{T}_t}$ and $P_{\mathcal{N}_t}$ accordingly. Since the generic mode-guidance vector $g^t_{\mathrm{mode}}(x_t;c_s)$ may possess components orthogonal to the manifold, we perform a manifold correction by subtracting its normal projection. This ensures that guidance remains tangent-aligned, thereby preserving both semantic consistency and geometric validity during the diffusion process.

A detailed algorithmic explanation of the entire process can be seen in Supplementary \cref{sec:additional_implementation}.

\section{Experimental Results}
\label{sec:results}

\begin{figure*}[!t]
    \centering
    \includegraphics[width=\linewidth, keepaspectratio]{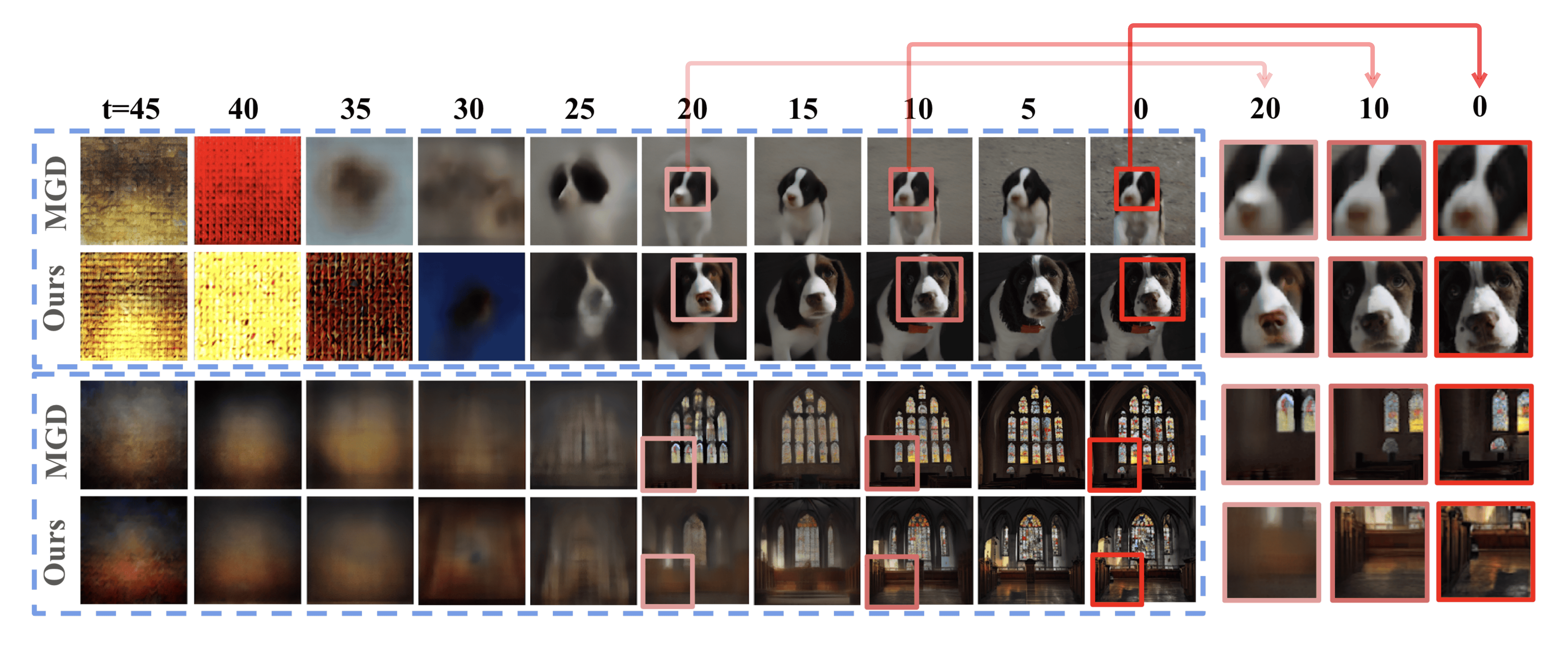}
    \vspace{-2.0em}
    \caption{\footnotesize
    \textbf{Qualitative evolution of generated samples across denoising timesteps ($t$).}
    Comparison between MGD~\cite{chan2025mgd3} and our \textbf{ManifoldGD} shows that early timesteps ($t{>}25$) capture coarse semantic structure via mode guidance, while later stages ($t{\leq}20$) refine geometry and texture under manifold constraints. ManifoldGD consistently yields sharper, more coherent generations with enhanced semantic fidelity and contrast. The zoomed insets (red boxes) highlight that MGD \cite{chan2025mgd3} often blurs key regions (e.g., missing eyes and unclear nose in the dog, coarse floor textures in the church), whereas ManifoldGD preserves fine-grained (e.g., different reflections from chairs and colored windows) lighting variations, reflections, and high-frequency details, demonstrating geometrically consistent and visually realistic synthesis.
    }
    \label{fig:per_step}
    \vspace{-1.5em}
\end{figure*}

\begin{figure}[!ht]
    \centering
    \includegraphics[width=\linewidth, keepaspectratio]{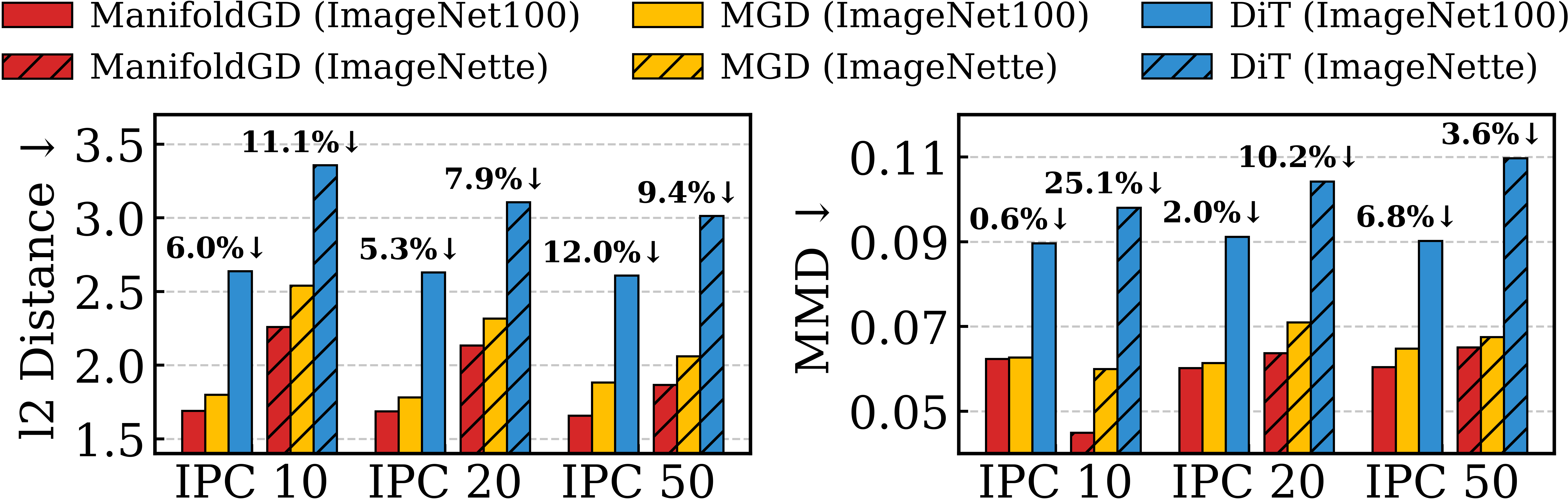}
    \caption{\footnotesize\textbf{$\ell_2$ and MMD comparison.} DiT \cite{peebles2023dit} achieves inferior results ($\uparrow$ $\ell_2$ MMD) and while ManifoldGD achieves the best result ($\downarrow$ L2 MMD).} 
    \label{fig:l2_mmd}
    \vspace{-1.5em}
\end{figure}

\textbf{Setup.} We evaluate our method using the hard-label protocol \cite{chan2025mgd3,gu2024minimaxdiff,su2024d4m}, the most challenging and unbiased setting for dataset distillation. Unlike soft-label or feature-matching variants that leak teacher information \cite{behrens2025dataset}, this setup trains student networks solely on $\mathcal{S}$ and discrete labels, ensuring improvements reflect true data fidelity. Experiments are conducted at image resolution $256\times256$ using ConvNet-6, ResNetAP-10, and ResNet-18 classifiers, with IPC = 10, 20, and 50 (10 being the most challenging due to least number of samples per class). We distill datasets for ImageNette \cite{howard2019imagenette}, ImageWoof \cite{howard2019imagewoof}, and ImageNet-100 \cite{tian2020contrastive}, and train classifiers from scratch for fair comparison. %All experiments run on NVIDIA RTX A6000 GPUs (48 GB) with mixed-precision inference and batch size 1 per diffusion trajectory.
Each setting is repeated three times with different seeds, and mean accuracy is reported. \textbf{Baselines.} We follow relevant baselines in \cref{sec:related} for comparison.

\noindent
\textbf{Metrics.} Following prior works~\cite{chan2025mgd3,gu2024minimaxdiff,su2024d4m}, we assess both downstream performance and distributional fidelity using five complementary metrics: \textbf{Classification Accuracy.}
We train the classifiers (ConvNet-6, ResNetAP-10, and ResNet-18) from scratch on $\mathcal{S}$ and test on  $\mathcal{D}_{\text{test}}$. $\mathrm{Acc}_{\mathcal{S}\to\mathcal{D}}$ reflects task fidelity and the semantic validity of the distilled samples. \textbf{Fréchet Inception Distance (FID).} FID~\cite{heusel2017gans} is computed between Inception-V3 embeddings of real ($\mathcal{D}_{\text{train}}$) and synthetic images ($\mathcal{S}$) to quantify global distribution alignment and perceptual realism. Lower values indicate higher visual fidelity. \textbf{$\ell_2$ and MMD.} We compute feature-space $\ell_2$ distance and Maximum Mean Discrepancy (MMD) between ResNet-18 embeddings (pretrained on ImageNet1k) of $\mathcal{D}_{\text{test}}$ and $\mathcal{S}$, capturing local and global manifold alignment respectively \cite{shen2024data,roy2025exchangeability}. Smaller values imply better consistency of $\mathcal{S}$ with $\mathcal{D}_{\text{test}}$. \textbf{Representativeness and Diversity.} To measure coverage and spread, we use cosine-based minimax scores computed from ResNet-18 (pretrained on ImageNet1k) embeddings: $\text{Rep} = \min_i \max_j \cos(f_{\mathcal{S}}^i, f_{\mathcal{D_{\text{train}}}}^j)$, $\text{Div} = 1 - \max_{i\neq j} \cos(f_{\mathcal{S}}^i, f_{\mathcal{S}}^j)$, $\cos(,)$ represents cosine similarity.
Higher \text{Representativeness} indicates better coverage of real data and higher \text{Diversity} reflects reduced redundancy.

\subsection{Comparison with Existing Methods}
\label{sec:sota}

\begin{table*}[t]
\centering
\
\caption{\footnotesize\textbf{Performance comparison with state-of-the-art methods on ImageNet subsets, evaluated using the hard-label protocol.} Results are based on ResNetAP-10 with average pooling, with the best performance highlighted in \textbf{bold}. \underline{Underlined} results are the one where ManifoldGD outperforms training-based methods. $\mathrm{Acc}_{\mathcal{S}\to\mathcal{D}}$ is used as the evaluation metric. * are the methods we re-implemented. Full indicates training the classifiers on the entire real training data and reporting classification accuracy on the real test data ($\mathrm{Acc}_{\mathcal{D_{\text{train}}}\to\mathcal{D_{\text{test}}}}$)}
\label{tab:comparison}
\resizebox{\linewidth}{!}{%
\begin{tabular}{@{}l|c|c|ccccccc|ccccc@{}}
\toprule
\multirow{2}{*}{\textbf{Dataset}} & \multirow{2}{*}{\textbf{IPC}} & \multirow{2}{*}{\textbf{Full}} & \multicolumn{7}{c|}{\textbf{Training-Based}} & \multicolumn{5}{c}{\textbf{Training-Free}} \\
\cmidrule(lr){4-10} \cmidrule(lr){11-15}
& & & \textbf{Herding} & \textbf{IDC-1} & \textbf{D$^{4}$M} & \textbf{DM} & \textbf{Li et al.} & \textbf{MinMaxDiff} & \textbf{Zou et al.} & \textbf{Random} & \textbf{LDM} & \textbf{DiT*} & \textbf{MGD*} & \underline{\textbf{ManifoldGD}} \\
& & & \cite{DBLP:conf/icml/Welling09} & \cite{pmlr-v162-kim22c} & \cite{su2024d4m} & \cite{zhao2023dm} & \cite{li2025task} & \cite{gu2024minimaxdiff} & \cite{zou2025dataset} & \cite{baseline-random} & \cite{rombach2022ldm} & \cite{peebles2023dit} & \cite{chan2025mgd3} & \\
\midrule
\multirow{3}{*}[0.5em]{\textbf{Nette}} 
& 10 & 95.1{\tiny$\pm$}1.2 & - & - & 60.4{\tiny$\pm$}3.4 & 60.8{\tiny$\pm$}0.6 & 61.5{\tiny$\pm$}0.9 & 62.0{\tiny$\pm$}0.2 & \textbf{64.8{\tiny$\pm$}3.6} & 54.2{\tiny$\pm$}1.6 & 60.3{\tiny$\pm$}3.6 & 59.1{\tiny$\pm$}0.7 & 61.9{\tiny$\pm$}0.6 & \textbf{64.1{\tiny$\pm$}1.2} \textcolor{green}{(+2.2)} \\
& 20 & 95.1{\tiny$\pm$}1.2 & - & - & 65.5{\tiny$\pm$}1.2 & 66.5{\tiny$\pm$}1.1 & 66.9{\tiny$\pm$}0.5 & 66.8{\tiny$\pm$}0.4 & \textbf{71.0{\tiny$\pm$}0.5} & 63.5{\tiny$\pm$}0.5 & 62.0{\tiny$\pm$}2.6 & 64.8{\tiny$\pm$}1.2 & 66.5{\tiny$\pm$}0.5 & \textbf{69.7{\tiny$\pm$}0.4} \textcolor{green}{(+3.2)} \\
& 50 & 95.1{\tiny$\pm$}1.2 & - & - & 73.8{\tiny$\pm$}1.7 & 76.2{\tiny$\pm$}0.4 & 76.8{\tiny$\pm$}0.7 & 76.6{\tiny$\pm$}0.2 & \textbf{81.2{\tiny$\pm$}0.8} & 76.1{\tiny$\pm$}1.1 & 71.0{\tiny$\pm$}1.4 & 73.3{\tiny$\pm$}0.9 & 77.5{\tiny$\pm$}1.1 & \textbf{78.4{\tiny$\pm$}1.0} \textcolor{green}{(+1.4)} \\
\midrule
\multirow{2}{*}[0.5em]{\textbf{ImageNet100}} 
& 10 & 80.3{\tiny$\pm$}0.2 & 19.8{\tiny$\pm$}0.3 & \textbf{25.7{\tiny$\pm$}0.1} & - & - & - & 24.8{\tiny$\pm$}0.2 & - & 19.1{\tiny$\pm$}0.4 & - & 23.2{\tiny$\pm$}0.3 & 26.1{\tiny$\pm$}0.6 & \underline{\textbf{27.6{\tiny$\pm$}0.5}} \textcolor{green}{(+1.5)} \\
& 20 & 80.3{\tiny$\pm$}0.2 & 27.6{\tiny$\pm$}0.1 & 29.9{\tiny$\pm$}0.2 & - & - & - & \textbf{32.3{\tiny$\pm$}0.1} & - & 26.7{\tiny$\pm$}0.5 & - & 28.4{\tiny$\pm$}0.4 & 33.2{\tiny$\pm$}0.5 & \underline{\textbf{35.3{\tiny$\pm$}0.5}} \textcolor{green}{(+2.1)} \\
\bottomrule
\end{tabular}%
}
\end{table*}

\noindent
\textbf{ImageNette and ImageNet-100:} As seen in \cref{tab:comparison}, ManifoldGD outperforms the existing training free data distillation methods for all IPCs. Even, compared to the training-based methods, we see comparable (in some cases even superior) performance demonstrating the effectiveness of the course correction of $g^t_{\text{manifold}}$. \textbf{ImageWoof:} Similar to ImageNette and ImageNet-100, \cref{tab:imagewoof} shows that ManifoldGD performs better than existing techniques.
The common expected trend of increase in $\mathrm{Acc}_{\mathcal{S}\to\mathcal{D}}$ with increase in number of samples (IPC) is seen across all datasets for all classifiers (\cref{tab:comparison} and \cref{tab:imagewoof}).

\begin{table*}[t]
\centering
\caption{\footnotesize\textbf{Performance comparison with state-of-the-art methods on ImageWoof, evaluated using the hard-label protocol.} Results are based on ResNetAP-10, ConvNet-6 and ResNet-18 with average pooling, with the best performance highlighted in \textbf{bold}. \underline{Underlined} results are the one where ManifoldGD outperforms training-based methods. $\mathrm{Acc}_{\mathcal{S}\to\mathcal{D}}$ is used as the evaluation metric. * are the methods we re-implemented. Full indicates training the classifiers on the entire real training data and reporting classification accuracy on the real test data ($\mathrm{Acc}_{\mathcal{D_{\text{train}}}\to\mathcal{D_{\text{test}}}}$)}
\resizebox{\linewidth}{!}{%
\begin{tabular}{l|l|c|ccccccc|cccc}
\toprule
\textbf{IPC (Ratio)} & \textbf{Test Model} 
& \textbf{Full}
& \multicolumn{7}{c|}{\textbf{Training-Based}} 
& \multicolumn{4}{c}{\textbf{Training-Free}} \\
& & \textbf{-}
& \textbf{Herding} & \textbf{DM} & \textbf{K-Center} & \textbf{Zou et al.} & \textbf{IDC-1} & \textbf{GLaD} & \textbf{MinMaxDiff}
  & \textbf{Random} & \textbf{DiT*} & \textbf{MGD*} & \underline{\textbf{ManifoldGD}} \\
& & 
& \cite{DBLP:conf/icml/Welling09} & \cite{zhao2023dm} & \cite{sener2018coreset} & \cite{zou2025dataset} & \cite{pmlr-v162-kim22c} & \cite{cazenavette2023glad} & \cite{gu2024minimaxdiff}
  & \cite{baseline-random} & \cite{peebles2023dit} & \cite{chan2025mgd3}
 & \\
\midrule

\multirow{3}{*}{\textbf{10 (0.8\%)}} 
& ConvNet-6   & 86.4{\tiny$\pm$0.2} & 26.7{\tiny$\pm$0.5} & 26.9{\tiny$\pm$1.2} & 19.4{\tiny$\pm$0.9} & 34.8{\tiny$\pm$2.4} & 33.3{\tiny$\pm$1.1} & 33.8{\tiny$\pm$0.9} & \textbf{37.0{\tiny$\pm$1.0}} & 24.3{\tiny$\pm$1.1} & 34.2{\tiny$\pm$1.1} & 35.1{\tiny$\pm$1.0} & \textbf{36.9{\tiny$\pm$0.6}} \textcolor{green}{(+1.8)} \\
& ResNetAP-10 & 87.5{\tiny$\pm$0.5} & 32.0{\tiny$\pm$0.3} & 30.3{\tiny$\pm$1.2} & 22.1{\tiny$\pm$0.1} & \textbf{39.5{\tiny$\pm$1.5}} & 39.1{\tiny$\pm$0.5} & 32.9{\tiny$\pm$0.9} & 39.2{\tiny$\pm$1.3} & 29.4{\tiny$\pm$0.8} & 34.7{\tiny$\pm$0.5} & 37.5{\tiny$\pm$2.2} & \textbf{38.3{\tiny$\pm$0.4}} \textcolor{green}{(+1.3)} \\
& ResNet-18   & 89.3{\tiny$\pm$1.2} & 30.2{\tiny$\pm$1.2} & 33.4{\tiny$\pm$0.7} & 21.1{\tiny$\pm$0.4} & \textbf{39.9{\tiny$\pm$2.6}} & 37.3{\tiny$\pm$0.2} & 31.7{\tiny$\pm$0.8} & 37.6{\tiny$\pm$0.9} & 27.7{\tiny$\pm$0.9} & 34.7{\tiny$\pm$0.4} & 37.9{\tiny$\pm$0.7} & \textbf{39.2{\tiny$\pm$0.7}} \textcolor{green}{(+1.3)} \\
\midrule
\multirow{3}{*}{\textbf{20 (1.6\%)}} 
& ConvNet-6   & 86.4{\tiny$\pm$0.2} & 29.5{\tiny$\pm$0.3} & 29.9{\tiny$\pm$1.0} & 21.5{\tiny$\pm$0.8} & \textbf{37.9{\tiny$\pm$1.9}} & 35.5{\tiny$\pm$0.8} & -- & 37.6{\tiny$\pm$0.2} & 29.1{\tiny$\pm$0.7} & 36.1{\tiny$\pm$0.8} & 36.5{\tiny$\pm$1.0} & \textbf{37.7{\tiny$\pm$0.6}} \textcolor{green}{(+1.2)} \\
& ResNetAP-10 & 87.5{\tiny$\pm$0.5} & 34.9{\tiny$\pm$0.1} & 35.2{\tiny$\pm$0.6} & 25.1{\tiny$\pm$0.7} & 44.5{\tiny$\pm$2.2} & 43.4{\tiny$\pm$0.3} & -- & \textbf{45.8{\tiny$\pm$0.5}} & 32.7{\tiny$\pm$0.4} & 41.1{\tiny$\pm$0.8} & 44.7{\tiny$\pm$1.5} & \textbf{45.6{\tiny$\pm$0.5}} \textcolor{green}{(+0.9)} \\
& ResNet-18   & 89.3{\tiny$\pm$1.2} & 32.2{\tiny$\pm$0.6} & 29.8{\tiny$\pm$1.7} & 23.6{\tiny$\pm$0.3} & \textbf{44.5{\tiny$\pm$2.0}} & 38.6{\tiny$\pm$0.2} & -- & 42.5{\tiny$\pm$0.6} & 29.7{\tiny$\pm$0.5} & 40.5{\tiny$\pm$0.5} & 40.9{\tiny$\pm$1.2} & \textbf{42.4{\tiny$\pm$0.4}} \textcolor{green}{(+1.4)} \\
\midrule
\multirow{3}{*}{\textbf{50 (3.8\%)}} 
& ConvNet-6   & 86.4{\tiny$\pm$0.2} & 40.3{\tiny$\pm$0.7} & 44.4{\tiny$\pm$1.0} & 36.5{\tiny$\pm$1.0} & \textbf{54.5{\tiny$\pm$0.6}} & 43.9{\tiny$\pm$1.2} & -- & 53.9{\tiny$\pm$0.6} & 41.3{\tiny$\pm$0.6} & 46.5{\tiny$\pm$0.8} & 50.8{\tiny$\pm$0.7} & \textbf{51.3{\tiny$\pm$0.2}} \textcolor{green}{(+0.5)} \\
& ResNetAP-10 & 87.5{\tiny$\pm$0.5} & 49.1{\tiny$\pm$0.7} & 47.1{\tiny$\pm$1.1} & 40.6{\tiny$\pm$0.4} & \textbf{57.3{\tiny$\pm$0.5}} & 48.3{\tiny$\pm$1.0} & -- & 56.3{\tiny$\pm$1.0} & 47.2{\tiny$\pm$1.3} & 49.3{\tiny$\pm$0.2} & 56.9{\tiny$\pm$0.6} & \underline{\textbf{57.9{\tiny$\pm$0.6}}} \textcolor{green}{(+1.0)} \\
& ResNet-18   & 89.3{\tiny$\pm$1.2} & 48.3{\tiny$\pm$1.2} & 46.2{\tiny$\pm$0.6} & 39.6{\tiny$\pm$1.0} & \textbf{58.9{\tiny$\pm$1.5}} & 48.3{\tiny$\pm$0.8} & -- & 57.1{\tiny$\pm$0.6} & 47.9{\tiny$\pm$1.8} & 50.1{\tiny$\pm$0.5} & 56.2{\tiny$\pm$0.4} & \textbf{58.2{\tiny$\pm$0.6}} \textcolor{green}{(+2.0)} \\
\midrule
\multirow{3}{*}{\textbf{70 (5.4\%)}} 
& ConvNet-6   & 86.4{\tiny$\pm$0.2} & 46.2{\tiny$\pm$0.6} & 47.5{\tiny$\pm$0.8} & 38.6{\tiny$\pm$0.7} & \textbf{55.8{\tiny$\pm$1.7}} & 48.9{\tiny$\pm$0.7} & -- & 55.7{\tiny$\pm$0.9} & 46.3{\tiny$\pm$0.6} & 50.1{\tiny$\pm$1.2} & 53.5{\tiny$\pm$0.5} & \textbf{55.3{\tiny$\pm$0.6}} \textcolor{green}{(+1.8)} \\
& ResNetAP-10 & 87.5{\tiny$\pm$0.5} & 53.4{\tiny$\pm$1.4} & 51.7{\tiny$\pm$0.8} & 45.9{\tiny$\pm$1.5} & \textbf{60.6{\tiny$\pm$0.3}} & 52.8{\tiny$\pm$1.8} & -- & 58.3{\tiny$\pm$0.2} & 50.8{\tiny$\pm$0.6} & 54.3{\tiny$\pm$0.9} & 58.8{\tiny$\pm$0.4} & \textbf{60.6{\tiny$\pm$0.1}} \textcolor{green}{(+1.8)} \\
& ResNet-18   & 89.3{\tiny$\pm$1.2} & 49.7{\tiny$\pm$0.8} & 51.9{\tiny$\pm$0.8} & 44.6{\tiny$\pm$1.1} & \textbf{60.3{\tiny$\pm$0.3}} & 51.1{\tiny$\pm$1.7} & -- & 58.8{\tiny$\pm$0.7} & 52.1{\tiny$\pm$1.0} & 51.5{\tiny$\pm$1.0} & 59.1{\tiny$\pm$0.6} & \underline{\textbf{61.0{\tiny$\pm$0.6}}} \textcolor{green}{(+1.9)} \\
\midrule
\multirow{3}{*}{\textbf{100 (7.7\%)}} 
& ConvNet-6   & 86.4{\tiny$\pm$0.2} & 54.4{\tiny$\pm$1.1} & 55.0{\tiny$\pm$1.3} & 45.1{\tiny$\pm$0.5} & \textbf{62.7{\tiny$\pm$1.4}} & 53.2{\tiny$\pm$0.9} & -- & 61.1{\tiny$\pm$0.7} & 52.2{\tiny$\pm$0.4} & 53.4{\tiny$\pm$0.3} & 59.2{\tiny$\pm$0.4} & \textbf{60.6{\tiny$\pm$0.2}} \textcolor{green}{(+1.4)} \\
& ResNetAP-10 & 87.5{\tiny$\pm$0.5} & 61.7{\tiny$\pm$0.9} & 56.4{\tiny$\pm$0.8} & 54.8{\tiny$\pm$0.2} & \textbf{65.7{\tiny$\pm$0.5}} & 56.1{\tiny$\pm$0.9} & -- & 64.5{\tiny$\pm$0.2} & 59.4{\tiny$\pm$1.0} & 58.3{\tiny$\pm$0.8} & 64.6{\tiny$\pm$0.1} & \textbf{65.2{\tiny$\pm$0.3}} \textcolor{green}{(+0.6)} \\
& ResNet-18   & 89.3{\tiny$\pm$1.2} & 59.3{\tiny$\pm$0.7} & 60.2{\tiny$\pm$1.0} & 50.4{\tiny$\pm$0.4} & \textbf{68.3{\tiny$\pm$0.4}} & 58.3{\tiny$\pm$1.2} & -- & 65.7{\tiny$\pm$0.4} & 61.5{\tiny$\pm$1.3} & 58.9{\tiny$\pm$1.3} & 65.5{\tiny$\pm$0.2} & \textbf{66.3{\tiny$\pm$0.6}} \textcolor{green}{(+0.8)} \\
\bottomrule
\end{tabular}
}
\label{tab:imagewoof}
\end{table*}

\begin{figure*}[!t]
    \centering
    \includegraphics[width=\linewidth, keepaspectratio]{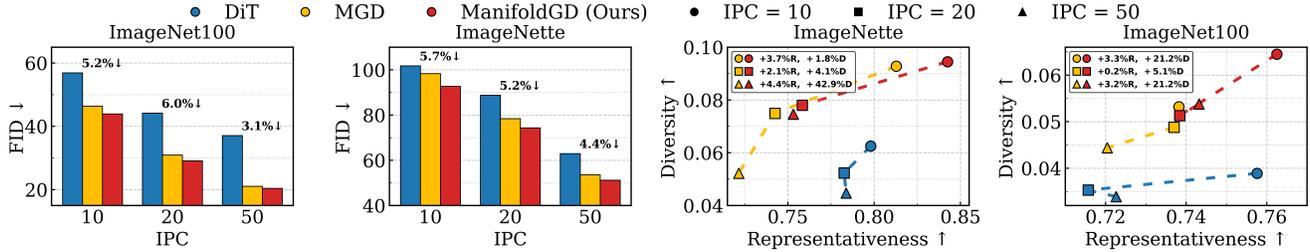}
    \vspace{-2.0em}\caption{\footnotesize\textbf{FID, Representativeness and Diversity comparison of DiT \cite{peebles2023dit}, MGD \cite{chan2025mgd3}, and ManifoldGD.} IPC 10,20, and 50 are used for ImageNet-100 and ImageNette. ManifoldGD achieves lower FID (\% drop over MGD  \cite{chan2025mgd3} marked above the bars), higher representativeness and diversity (R=representativeness, D=diversity, \% increase over MGD  \cite{chan2025mgd3} shown in the plot) across all settings.} 
    \label{fig:fid_rep_div_comparison}
    \vspace{-1.5em}
\end{figure*}

In addition to $\mathrm{Acc}_{\mathcal{S}\to\mathcal{D}}$, we evaluate FID across DiT \cite{peebles2023dit}, MGD \cite{chan2025mgd3}, and our ManifoldGD on ImageNette and ImageNet-100 with IPC = 10, 20, and 50. As shown in Figure \cref{fig:fid_rep_div_comparison}, DiT yields the highest (worst) FID, MGD improves both FID and accuracy, while ManifoldGD achieves the lowest FID and highest $\mathrm{Acc}_{\mathcal{S}\to\mathcal{D}}$ across all settings. This consistent trend supports the link between generative fidelity and downstream generalization: higher FID indicates synthetic shortcuts \cite{geirhos2020shortcut} that hinder real-data performance, while manifold-consistent guidance preserves semantic alignment and improves generalization. Moreover, ManifoldGD exhibits superior representativeness and diversity (\cref{fig:fid_rep_div_comparison}), confirming that hierarchical divisive clustering yields a compact yet expressive coreset that balances coarse semantic coverage with fine intra-class variability.

\cref{fig:l2_mmd} reports the $\ell_2$ and $\mathrm{MMD}$ distances between the synthetic and real distributions, serving as complementary measures of distribution-level fidelity between the $\mathcal{S}$ (train set) and $\mathcal{D_{\text{test}}}$ \cite{roy2025exchangeability, shen2024data, mehta2023efficient}. ManifoldGD consistently attains the lower $\ell_2$ and $\mathrm{MMD}$ values, indicating that its generated samples are both closer to individual real samples and better aligned with the overall data distribution.

We also see in Supplementary Section \cref{sec:additional_ablation} Table \cref{tab:ddim} that ManifoldGD outperforms MGD \cite{chan2025mgd3} while using DDIM scheduler as well, showing effectiveness irrespective of the scheduler configuration.

\subsection{Ablation Studies}
\label{sec:ablation}

\begin{figure}[!t]
    \centering
    \includegraphics[width=0.8\linewidth, keepaspectratio]{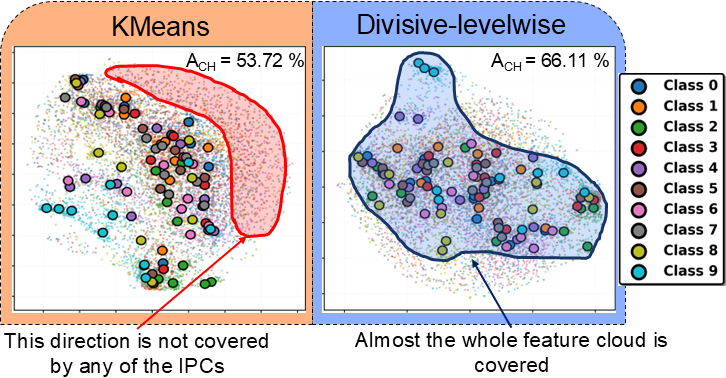}
    \caption{\footnotesize\textbf{VAE feature space and IPC illustration for Nette IPC=10.} KMeans do not capture the entire feature cloud whereas our divisive-levelwise occupy regions near the mean as well as the outlines of the feature cloud. The convex hull area ratio (in percentage) $A_{CH} = \frac{\text{Area}(\text{IPC entroid ($c_s$) hull})}{\text{Area}(\text{entire VAE feature space hull})}$ is higher for our divisive method, indicating better spatial coverage.} 
    \label{fig:feature_space}
    \vspace{-1.5em}
\end{figure}

\paragraph{Impact of Divisive-levelwise clustering.}
We analyze the proposed divisive level-wise IPC selection for capturing global and local semantic variations in VAE latent space. \cref{tab:ablation_clustering} shows our strategy consistently outperforms $k$-means and other hierarchical clustering baselines, while \cref{fig:feature_space} illustrates that our level-wise traversal spans the feature cloud more uniformly versus $k$-means' limited sub-regions. Agglomerative clustering performs worst, placing centroids near outer boundaries (Supplementary \cref{sec:additional_ablation}). Optimal $s_{\text{start}}$ depends on class granularity: deeper splits for well-separated classes, shallower splits for overlapping classes to avoid noisy leaf regions. \textit{\underline{Takeaway}:} Hierarchical clustering outperforms $k$-means via coarse-to-fine hierarchy, with top-down divisive selection being more appropriate than bottom-up agglomerative approaches. Level-wise centroid selection enhances coverage, with $s_{\text{start}}$ adapting to dataset class separation.

\begin{table}[t]
\centering
\caption{\footnotesize\textbf{Ablation study on the components of the proposed method.} All evaluations are done for IPC = 10 on ImageNette. C $\to$ Clustering, L $\to$ Level-wise, A $\to$ Annealing}
\vspace{-1.0em}
\resizebox{\linewidth}{!}{
\begin{tabular}{lccc|ccc}
\toprule
\textbf{Method} & \textbf{C} & \textbf{L} & \textbf{A} & \textbf{ConvNet-6} & \textbf{ResNet-AP-10} & \textbf{ResNet-18} \\
\midrule
KMeans & \cmark & \xmark & \xmark & 56.3 & 61.0 & 59.7 \\
\small{Agglomerative} & \cmark & \xmark & \xmark & 37.7 & 45.9 & 42.6 \\
Divisive & \cmark & \xmark & \xmark & 57.4 & 62.5 & 58.5 \\
\small{Divisive-levelwise} & \cmark & \cmark & \xmark & 59.2 & 63.3 & 61.1 \\
\hline
Ours & \cmark & \cmark & \xmark & 60.5 & 64.0 & 62.3 \\
Ours (annealed) & \cmark & \cmark & \cmark & 60.8 & 64.5 & 62.7 \\
\bottomrule
\end{tabular}
}
\vspace{-1.0em}
\label{tab:ablation_clustering}
\end{table}

\vspace{-0.2in}
\paragraph{Impact of $g^t_{\text{manifold}}$.}
\cref{tab:ablation_clustering} further highlights the effectiveness of the proposed manifold guidance ($g^t_{\text{manifold}}$), which refines the coarse semantic attraction of ($g^t_{\text{mode}}$) by re-projecting samples back onto the estimated tangent manifold. When combined with divisive-levelwise clustering, this correction consistently yields improved performance, validating that manifold-consistent attraction effectively complements hierarchical IPC centroid selection. Furthermore, \cref{tab:warm_schedule_backbones} analyzes the influence of different radius decay schedules for the local neighborhood ($\mathcal{N}_s$) used in tangent estimation. We find that an exponential annealing of the neighborhood radius achieves the best performance. This faster decay emphasizes progressively finer local structures of the manifold as diffusion proceeds, aligning with the intuition that higher-noise timesteps should rely on broader geometric context, while later steps should adapt to tighter, more locally linear approximations of ($\mathcal{M}_t$). \textit{\underline{Takeaway}:} The introduction of $g^t_{\text{manifold}}$ further enhances the performance achieved by applying hierarchical clustering. Annealing the radius of $\mathcal{N}_s$ operationalizes the trade-off in \cref{remark_1}: larger initial radii accommodate necessary exploration when manifold estimates are uncertain, while progressive tightening enforces geometric constraints as the manifold becomes better defined, preventing over smoothing while maintaining consistency.

\begin{table}[t]
\centering
\caption{\footnotesize\textbf{Comparison of learning rate schedules across warm-up steps.} 
Results are reported for ConvNet-6 and ResNet-18. Faster decay of the radius of $\mathcal{N_s}$ achieves the best result.}
\vspace{-1.0em}
\resizebox{\linewidth}{!}{
\begin{tabular}{c|cc|cc|cc}
\toprule
& \multicolumn{2}{c|}{\textbf{Exponential}} 
& \multicolumn{2}{c|}{\textbf{Cosine}} 
& \multicolumn{2}{c}{\textbf{Linear}} \\
\cmidrule(lr){2-3} \cmidrule(lr){4-5} \cmidrule(lr){6-7}
\textbf{t} 
& \textbf{ConvNet-6} & \textbf{ResNet-18} 
& \textbf{ConvNet-6} & \textbf{ResNet-18} 
& \textbf{ConvNet-6} & \textbf{ResNet-18} \\
\midrule
5  & 60.8 & 62.7 & 59.4 & 62.7 & 59.6 & 62.5 \\
10 & 60.1 & 61.9 & 58.3 & 62.1 & 58.1 & 61.2 \\
15 & 58.5 & 62.1 & 57.9 & 61.9 & 57.8 & 62.3 \\
\bottomrule
\end{tabular}
}
\label{tab:warm_schedule_backbones}
\vspace{-1.5em}
\end{table}

\begin{figure}[!t]
    \centering
    \includegraphics[width=\linewidth, keepaspectratio]{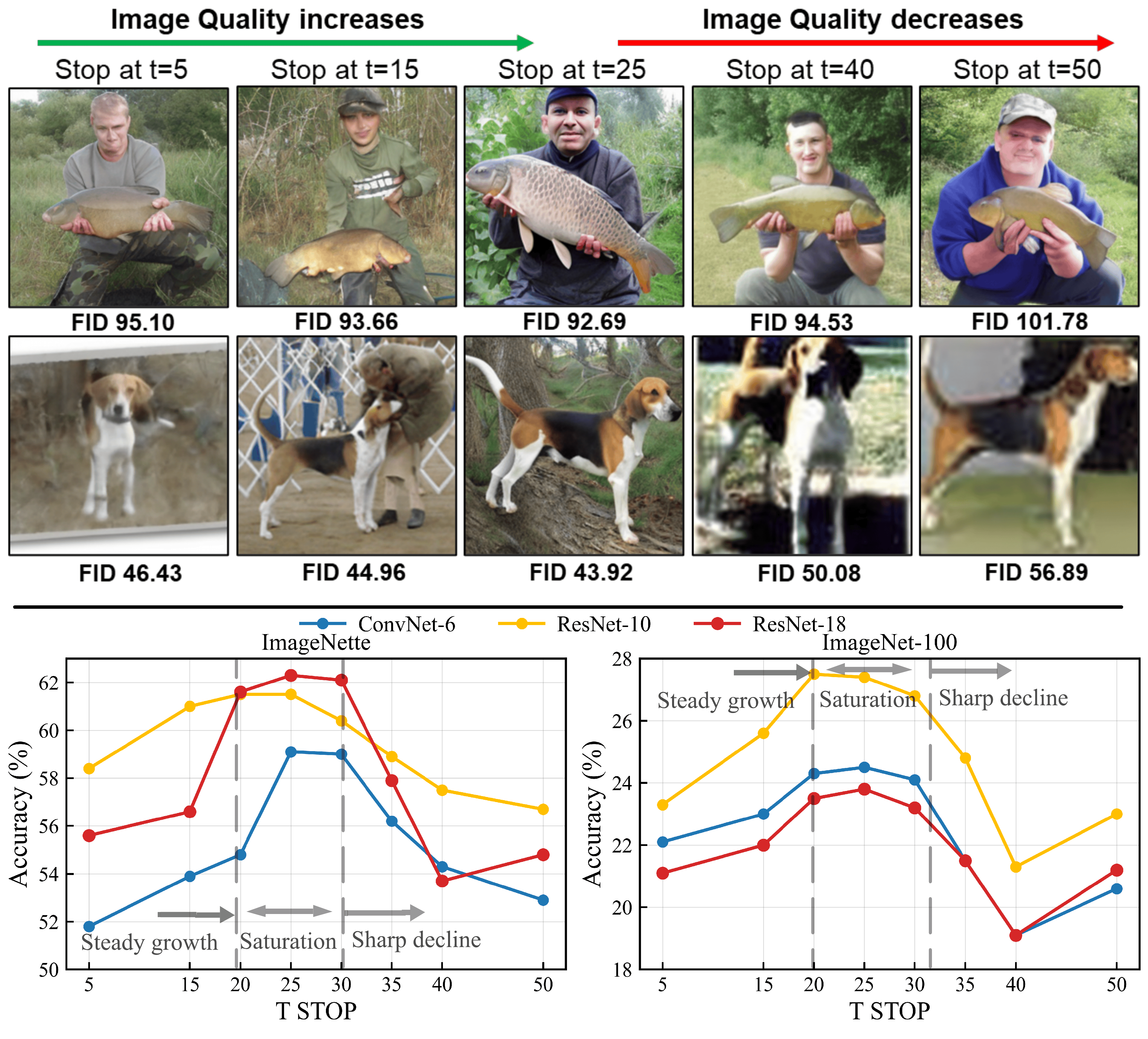}
    \vspace{-2.0em}
    \caption{\footnotesize\textbf{Ablation experiments varying $T_{STOP}$ for ImageNette (first row) and ImageNet-100 (last row).} Increasing $T_{STOP}$ till 20 (steady growth region) improves image quality ($\downarrow$ FID) and classification performance ($\uparrow$ $\mathrm{Acc}_{\mathcal{S}\to\mathcal{D}}$). From $T_{STOP}=20 \to 30$ we see a stable FID and $\mathrm{Acc}_{\mathcal{S}\to\mathcal{D}}$ (saturation region). Further increase in $T_{STOP}$ beyond 30 leads to drop in both image quality ($\uparrow$ FID) and performance ($\downarrow$ $\mathrm{Acc}_{\mathcal{S}\to\mathcal{D}}$).} 
    \label{fig:ab_stop}
    \vspace{-1.0em}
\end{figure}

\vspace{-0.2in}
\paragraph{Ablation on varying $T_{STOP}$.} \cref{fig:ab_stop} illustrates the effect of varying $T_{\text{STOP}}$ (the timestep at which guidance stops) on both visual fidelity and downstream accuracy. Increasing $T_{\text{STOP}}$ up to 25 (out of 50 total denoising steps) improves FID and classification performance, suggesting that extending guidance through the early-to-mid denoising stages enables sufficient manifold correction before entering the low-noise regime. However, beyond this point, both metrics deteriorate, implying that excessive guidance interferes with the natural score-based refinement, constraining the generation to overly restricted regions of the diffusion manifold. \textit{\underline{Takeaway}:} FID and $\mathrm{Acc}_{\mathcal{S}\to\mathcal{D}}$ increase steadily till $T_{STOP}$ = 25 and decreases rapidly beyond that point. Since most structural formation occurs in early timesteps, guidance during later, fine-detail stages leads to overfitting and hinder natural denoising dynamics \cite{chen2025influence}.

\begin{figure}[!t]
    \centering
    \includegraphics[width=\linewidth, keepaspectratio]{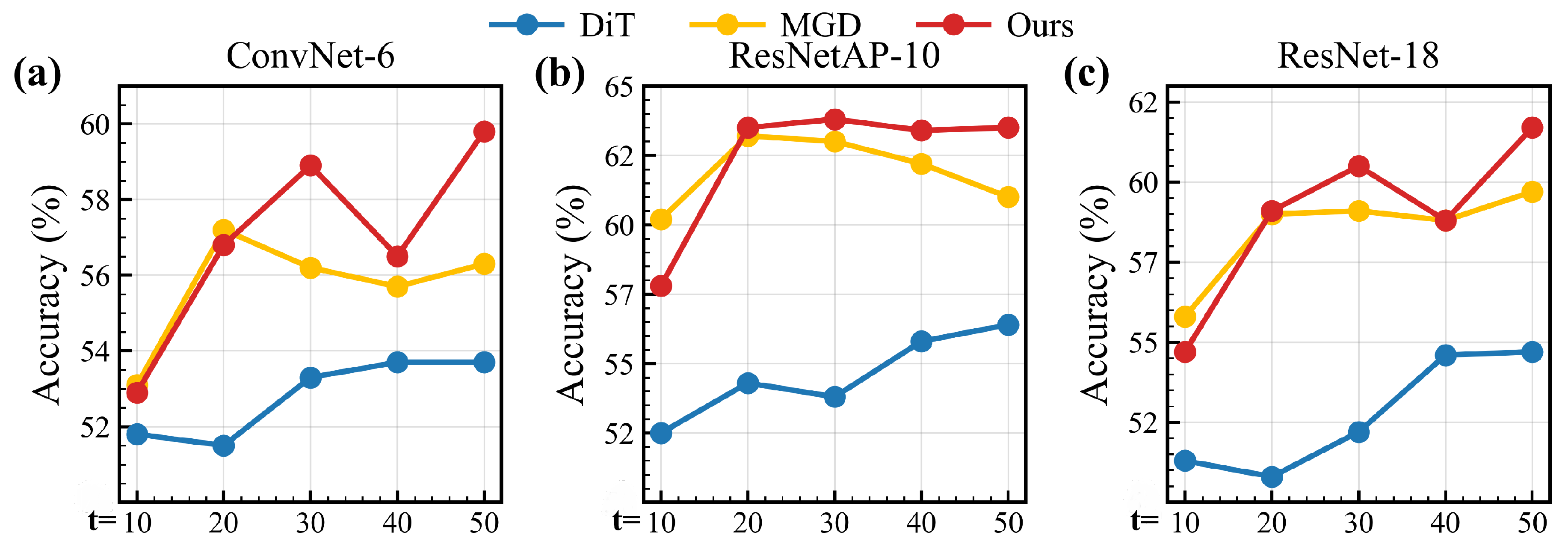}
    \vspace{-2.0em}
    \caption{\footnotesize \textbf{Ablation of diffusion step $t$ and corresponding $T_{\text{STOP}}$.} 
    We vary $t \in \{10, 20, 30, 40, 50\}$ and set $T_{\text{STOP}} = t/2$. The results show that mode guidance dominates the early denoising phase, while manifold guidance progressively takes over as $t$ increases. This transition indicates that early denoising benefits from strong class-alignment forces ($g^t_{mode}$), whereas later stages require geometric correction ($g^t_{manifold}$) to maintain fidelity and prevent off-manifold drift.}
    \label{fig:ablation_t_tstop}
    \vspace{-1.0em}
\end{figure}

\vspace{-0.2in}
\paragraph{Ablation on varying $t$.}
We investigate the interaction between denoising steps $t$ and $T_{\text{STOP}}$ in \cref{fig:ablation_t_tstop}. For smaller $t$, $g^t_{\text{mode}}$ dominates as $x_t$ lies far from the data manifold, requiring strong semantic alignment when manifold structure is weakly defined. As $t$ increases and samples approach high-density regions, $g^t_{\text{manifold}}$ becomes critical for refining details and preventing off-manifold drift. \textit{\underline{Takeaway}:} Better $\mathrm{Acc}{\mathcal{S}\to\mathcal{D}}$ favors $g^t_{\text{mode}}$ at lower $t$ and $g^t_{\text{manifold}}$ at higher $t$, directly addressing \cref{remark_1}'s balance: early mode guidance enables exploration, while later manifold correction ensures consistency without oversmoothing.

\vspace{-0.2in}
\paragraph{Ablation on varying $\phi$.} The results in \cref{tab:kernel_ablation} validate that $g^t_{\text{manifold}}$ is kernel-agnostic (see \cref{sec:mode_guidance}). While RBF kernel (corresponds to MGD \cite{chan2025mgd3}) provides a smooth isotropic attraction toward $c_s$, both Laplace (with sharper, local gradients) and IMQ (with heavier tails) exhibit comparable improvements when combined with our manifold correction term $g^t_{\text{manifold}}$. \textit{\underline{Takeaway}:} Consistent gain across ConvNet-6, ResNet-18, and ResNetAP-10 reinforces that $g^t_{\text{manifold}}$ provides a robust geometric correction independent of classifier capacity and is kernel-agnostic.

\begin{table}[t]
\centering
\caption{\footnotesize\textbf{Ablation on kernel functions for mode guidance on ImageNette (IPC = 10).} 
We compare RBF (MGD \cite{chan2025mgd3}), Laplace, and IMQ kernels with and without manifold guidance across three classifiers.
Manifold guidance consistently improves performance regardless of kernel type, demonstrating its general applicability.}
\vspace{-0.5em}
\resizebox{\linewidth}{!}{
\vspace{-0.5em}
\begin{tabular}{c|ccc|ccc}
\toprule
\multirow{2}{*}{\textbf{Kernel}} & \multicolumn{3}{c|}{\textbf{Without manifold guidance ($g^t_{mode}$)}} & \multicolumn{3}{c}{\textbf{\textbf{With manifold guidance ($g^t_{mode}$} + $g^t_{man}$})} \\
\cmidrule(lr){2-4}\cmidrule(lr){5-7}
 & ConvNet-6 & ResNet-18 & ResNetAP-10 & ConvNet-6 & ResNet-18 & ResNetAP-10 \\
\midrule
RBF       & 56.7 & 61.5 & 62.5 & \textbf{58.5} & \textbf{62.5} & \textbf{64.4} \\
Laplace   & 56.3 & 56.1 & 58.8 & \textbf{57.0} & \textbf{57.2} & \textbf{60.2} \\
IMQ       & 54.4 & 56.1 & 57.1 & \textbf{55.6} & \textbf{56.7} & \textbf{58.5} \\
\bottomrule
\end{tabular}
}
\vspace{-1.5em}
\label{tab:kernel_ablation}
\end{table}

\vspace{-0.2in}
\paragraph{Qualitative Analysis.}
\cref{fig:per_step} illustrates that ManifoldGD produces sharper and more stable generations than MGD \cite{chan2025mgd3} across denoising steps. At smaller $t$, the manifold correction constrains updates along the local tangent of the diffusion manifold $\mathcal{M}_t$, maintaining geometric consistency while preserving class-level alignment through mode guidance. This yields cleaner, high-contrast details (e.g., facial textures, structural edges) that persist throughout sampling. In contrast, MGD \cite{chan2025mgd3} shows visible texture blurring and geometric drift at later steps, indicating off-manifold deviations caused by unconstrained Euclidean updates. Similar trends appear in \cref{fig:qualitative} where ManifoldGD achieves higher quality and semantically sound images resembling the original samples more, whereas MGD \cite{chan2025mgd3} and DiT \cite{peebles2023dit} fail.

\vspace{-0.5em}
\section{Conclusion}
\label{sec:con}
We introduce Manifold-Guided Diffusion (ManifoldGD), a geometry-aware dataset distillation framework that unifies semantic mode attraction with manifold-consistent correction. By decomposing the conditional score into mode-aligned and geometry-constrained components, ManifoldGD ensures synthetic samples remain faithful to both class semantics and intrinsic data geometry. Leveraging tangent-normal decomposition on diffusion manifolds derived from hierarchical IPC centroid clustering, our approach preserves local structural validity while maintaining global diversity. Empirical results across multiple datasets, backbones, and kernel variants show consistent gains in classification accuracy, FID, and perceptual fidelity over prior methods, demonstrating that balancing semantic alignment and geometric consistency yields more representative distilled datasets.  
\noindent
\textbf{Limitations and future work.}  
At high noise levels, diffusion corrupts local neighborhoods, biasing tangent estimates and weakening manifold reconstruction. Similarly, low-rank approximations struggle with highly curved manifolds, risking over smoothing and limited diversity (\cref{remark_1}). We mitigate these via (i) adaptive $(r, K_t)$, (ii) ridge-regularized covariance, and (iii) annealing $\lambda_{\mathrm{man}}$ (\cref{sec:additional_implementation}) to balance consistency and diversity. While empirically stable (Sec.~\ref{sec:ablation}), formal analysis of projection error and curvature sensitivity remains future work.

\textbf{Acknowledgment}
Prof. Lokhande acknowledges support from University at Buffalo startup funds, an Adobe Research Gift, an NVIDIA Academic Grant, and the National Center for Advancing Translational Sciences of the NIH (award UM1TR005296 to the University at Buffalo). Dr.\ Chakraborty performed this work under the auspices of the U.S.\ Department of Energy by Lawrence Livermore National Laboratory under Contract DE-AC52-07NA27344.
{
    \small
    \bibliographystyle{ieeenat_fullname}
    \bibliography{main}
}

\clearpage
\setcounter{page}{1}
\maketitlesupplementary

\section{Algorithmic explanation of ManifoldGD}
\label{sec:additional_implementation}

\begin{table*}[!t]
\centering
\small
\caption{\textbf{Class-wise Top-1 Accuracy (\%) on ImageWoof for IPC=10 and IPC=50.}
MGD~\cite{chan2025mgd3} vs ManifoldGD across three classifiers. Columns list the 10 ImageWoof classes. Class abbreviations: Aust.\ Terrier (Australian Terrier), Border Terrier, Samoyed, Beagle, Shih-Tzu, Rhodesian Ridgeback, Dingo, Golden Retriever, Old English Sheepdog, German Shepherd.}
\label{tab:imagewoof_classwise_rotated}
\resizebox{\linewidth}{!}{
\begin{tabular}{ll|cccccccccc}
\toprule
\textbf{Setting} & \textbf{Method} &
\textbf{Aust.\ Terr.} & 
\textbf{Border Terr.} & 
\textbf{Samoyed} & 
\textbf{Beagle} & 
\textbf{Shih-Tzu} &
\textbf{Ridgeback} & 
\textbf{Dingo} & 
\textbf{Golden Ret.} &
\textbf{Old Eng.\ Sheep.} &
\textbf{Germ.\ Shep.}
\\ \midrule
\multicolumn{12}{c}{\textbf{IPC = 10}} \\
\midrule
\multirow{2}{*}{ConvNet} & MGD \cite{chan2025mgd3} & 24.9 & 40.7 & 16.7 & 27.7 & 22.2 & 31.2 & 28.9 & 31.8 & 40.3 & 32.2 \\
& ManifoldGD & 24.7 & 36.8 & 21.5 & 40.6 & 31.2 & 36.1 & 28.7 & 34.6 & 34.5 & 35.6 \\
\midrule
\multirow{2}{*}{ResNet-AP} & MGD \cite{chan2025mgd3} & 35.0 & 44.1 & 21.5 & 46.0 & 24.7 & 32.4 & 29.4 & 37.4 & 38.9 & 45.9 \\
& ManifoldGD & 36.4 & 39.3 & 22.6 & 49.1 & 35.2 & 39.6 & 38.4 & 48.6 & 38.2 & 50.0 \\
\midrule
\multirow{2}{*}{ResNet-18} & MGD \cite{chan2025mgd3} & 28.9 & 43.4 & 18.2 & 37.5 & 37.2 & 31.7 & 36.2 & 34.4 & 45.2 & 35.6 \\
& ManifoldGD & 34.7 & 43.5 & 22.7 & 44.2 & 34.2 & 41.5 & 35.4 & 46.9 & 41.7 & 36.8 \\
\midrule
\multicolumn{12}{c}{\textbf{IPC = 50}} \\
\midrule
\multirow{2}{*}{ConvNet} & MGD \cite{chan2025mgd3} & 47.4 & 58.3 & 39.2 & 50.4 & 54.9 & 45.5 & 43.9 & 50.7 & 53.6 & 56.8 \\
& ManifoldGD & 48.7 & 54.4 & 36.6 & 53.1 & 61.8 & 52.1 & 49.6 & 50.9 & 60.4 & 58.3 \\
\midrule
\multirow{2}{*}{ResNet-AP} & MGD \cite{chan2025mgd3} & 57.0 & 69.4 & 50.2 & 52.7 & 57.6 & 52.1 & 55.1 & 49.8 & 63.4 & 56.1 \\
& ManifoldGD & 54.0 & 66.0 & 53.5 & 59.4 & 59.1 & 58.0 & 61.1 & 55.5 & 62.7 & 64.9 \\
\midrule
\multirow{2}{*}{ResNet-18} & MGD \cite{chan2025mgd3} & 50.6 & 66.9 & 44.0 & 47.3 & 61.1 & 54.3 & 55.1 & 54.0 & 63.2 & 60.5 \\
& ManifoldGD & 53.3 & 63.7 & 42.5 & 61.2 & 64.3 & 60.2 & 57.1 & 58.1 & 62.9 & 61.2 \\
\bottomrule
\end{tabular}}
\end{table*}

Given an IPC centroid $c_s$ and its static latent neighborhood $\mathcal N_s$ (A detailed algorithmic explanation of the formation of $\mathcal N_s$ can be seen in \cref{alg:hier_ipc_static}.), the time-aligned manifold patch at diffusion timestep $t$ is obtained by forward-diffusing every $z \in \mathcal N_s$ using the DDPM noise schedule $\varepsilon_t^{(k)} \sim \mathcal N(0,(1-\bar\alpha_t)I)$:
\[
\mathcal M_t^{(s)} = 
\Big\{\, x_t^{(k)} = \sqrt{\bar\alpha_t}\, z_k + \varepsilon_t^{(k)} 
\;\Big|\; 
z_k \in \mathcal N_s \Big\}
\]
Here the centroid $c_s$ is fixed per synthetic sample: every reverse trajectory is conditioned on the specific $c_s$ assigned at initialization.

At each timestep we compute local geometry relative to the current latent $x_t$ by taking its $K_t$ nearest neighbors ($K_t=300$ empirically works the best) \emph{inside its own} time-aligned patch $\mathcal M_t^{(s)}$:
\[
\mathcal N_t = \text{KNN}(x_t;\,\mathcal M_t^{(s)}, K_t).
\]
This produces a time-dependent covariance
\[
C_t = \frac{1}{|\mathcal N_t|}\sum_{x\in\mathcal N_t} (x-\bar{x})(x-\bar{x})^\top,
\]
whose top-$d$ ($d=3$ empirically works the best) eigenvectors define the tangent and normal projectors,
\[
P_{\mathcal T_t} = U_{1:d}U_{1:d}^\top, \qquad
P_{\mathcal N_t} = I - P_{\mathcal T_t}.
\]
The manifold-corrected guidance is obtained by canceling the normal component of the mode-guidance vector,
\[
g^{t}_{\mathrm{manifold}}(x_t;c_s)
=
-\,P_{\mathcal N_t}\,g^t_{\mathrm{mode}}(x_t;c_s),
\]
thereby restricting conditional attraction to the tangent directions of the evolving class manifold and preventing off-manifold drift.

\begin{algorithm}[t]
\small
\caption{Hierarchical IPC Selection \& Static Neighborhoods}
\label{alg:hier_ipc_static}
\begin{algorithmic}[1]
\Require Latents $Z=\{z_i\}_{i=1}^N$, IPC budget $K$, max depth $L$, start level $s_{\mathrm{start}}$, radius $r$, tangent dim $d$, ridge $\gamma$
\Ensure Selected centroids $\{c_s,\mathcal N_s\}_{s=1}^K$
\vspace{0.3em}
\State Build divisive binary tree (bisecting $k$-means, SSE split) until depth $\le L$
\State Let $\mathcal L_d$ be leaf nodes at depth $d$ for $d=0,\dots,L$
\State $\mathcal S \leftarrow \varnothing$, $k \leftarrow 0$
\Comment{Stage 1: repeated coarse$\to$fine rounds}
\While{$k < K$}
  \For{$d = s_{\mathrm{start}}$ \textbf{downto} $0$}
    \If{$k < K$ and $\mathcal L_d \neq \varnothing$}
      \State sample $n \sim \mathcal L_d$ uniformly; $\mathcal S \leftarrow \mathcal S \cup \{n\}$; remove $n$ from $\mathcal L_d$
      \State $k \leftarrow k + 1$
    \EndIf
  \EndFor
  \State $s_{\mathrm{start}} \leftarrow s_{\mathrm{start}} + 1$ \Comment{expand sweep outward}
  \If{$s_{\mathrm{start}} > L$} \textbf{break} \EndIf
\EndWhile
\Comment{Stage 2: deep-fill remaining quota}
\If{$k < K$}
  \State $\mathcal R \leftarrow \bigcup_{d=0}^L \mathcal L_d$; sample $(K-k)$ nodes uniformly from $\mathcal R$; add to $\mathcal S$
  \State $k \leftarrow K$
\EndIf
\vspace{0.3em}
\Comment{Static neighborhoods (computed offline)}
\For{each selected node $n_s \in \mathcal S$}
  \State $c_s \leftarrow$ centroid of $n_s$ (mean or medoid)
  \State $\mathcal N_s \leftarrow \{ z\in Z : \|z - c_s\|_2 \le r \}$ \Comment{static latent neighborhood}
  % \If{$|\mathcal N_s| = 0$} \State $\mathcal N_s \leftarrow$ nearest-$1$ sample to $c_s$ in $Z$ \EndIf
\EndFor
\State \Return $\{c_s,\mathcal N_s\}_{s=1}^K$
\end{algorithmic}
\normalsize
\end{algorithm}

\section{Baselines used for comparison}
\label{sec:additional_baselines}

We categorize existing dataset distillation methods into three primary groups: i) traditional coreset selection, ii) training-based diffusion methods, and iii) training-free diffusion methods. This categorization provides a structured landscape against which we position ManifoldGD. i)\underline{Coreset selection:} Coreset selection techniques, such as \textbf{Herding} \cite{DBLP:conf/icml/Welling09} and \textbf{K-Center} \cite{sener2018coreset}, aim to identify a representative subset of the original data that captures the underlying data distribution. While efficient, these methods often struggle to capture the full complexity of the data distribution and exhibit limited cross-architecture generalization. ii)\underline{Training-based diffusion methods:} Training-based methods fine-tune or optimize within the generative model to produce synthetic datasets. \textbf{DM} \cite{zhao2023dm} utilizes a latent diffusion model to distill datasets, showing strong generalization. \textbf{Min-Max Diffusion} \cite{gu2024minimaxdiff} employs a min-max objective to enforce diversity and representativeness, though it requires generator optimization. Methods like \textbf{D$^{4}$M} \cite{su2024d4m} and \textbf{Zou et al.} \cite{zou2025dataset} also fall into this category, achieving high performance but incurring the computational cost associated with model training or fine-tuning. \textbf{IDC-1} \cite{pmlr-v162-kim22c} and \textbf{GLAD} \cite{cazenavette2023glad} are also training-based methods that optimize synthetic images through gradient matching and generative modeling respectively. iii)\underline{Training-free diffusion methods:} \textbf{Random} \cite{baseline-random} simply selects random real images, providing a lower-bound performance benchmark. \textbf{Latent Diffusion Models (LDM)} \cite{rombach2022ldm} and \textbf{DiT} \cite{peebles2023dit} can be directly sampled from, but without guidance, the resulting datasets may lack semantic focus. \textbf{MGD} \cite{chan2025mgd3} proposes a mode-guided diffusion pipeline that identifies class prototypes via clustering and guides the sampling process towards them, significantly improving upon unguided sampling. 

Several concurrent works have explored enhanced guidance strategies for diffusion-based dataset distillation. \cite{ye2025information} employs an information-theoretic framework that requires a separately trained classifier to guide the sampling process. \cite{chen2025influence} utilizes a pre-trained classifier to compute influence functions for guiding the generation trajectory. Similarly, other methods leverage text encoders for enhanced semantic conditioning \cite{zou2025dataset}, additional classifiers for OOD detection \cite{gao2025good}, or other external guidance mechanisms. Unlike these approaches that rely on extra components beyond the diffusion model itself, our ManifoldGD framework operates with only a single pre-trained diffusion model and a VAE for feature extraction. To maintain a fair comparison, we do not directly compete with these more complex architectures in our main experiments. Nevertheless, as seen in Section \cref{sec:sota}, our method achieves comparable performance to the results reported in these papers.

\section{Ablation Experiments}
\label{sec:additional_ablation}

\begin{figure}[!t]
    \centering
    \includegraphics[width=\linewidth, keepaspectratio]{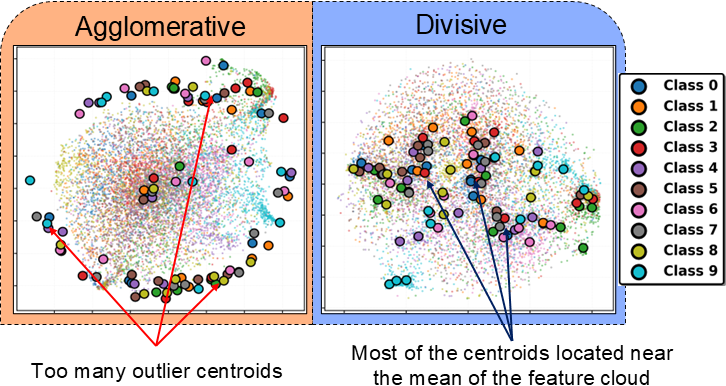}
    \caption{\footnotesize\textbf{VAE feature space and IPC illustration for Nette IPC=10.} Agglomerative clustering outputs IPC cluster centroids that near the edge of the feature cloud whereas Divisive clustering outputs IPC centroids that are near the mean of the feature cloud.} 
    \label{fig:ab_agglo}
\end{figure}

\noindent
\textbf{Agglomerative vs. divisive clustering for IPC centroid selection.} \cref{fig:ab_agglo} highlights a key structural difference between agglomerative clustering and divisive (bisecting) clustering when operating in the VAE latent space. Agglomerative clustering begins with individual samples and progressively merges them ("bottom up" approach), causing early merges to be dominated by peripheral or low-density points. As a result, the resulting IPC centroids tend to lie near the boundary of the feature cloud, often reflecting outlier- or noise-driven directions. In contrast, divisive clustering recursively splits clusters, producing progressively purer, high-density partitions. This top-down refinement naturally places IPC centroids closer to the mean or high-density core of the feature manifold, yielding more stable and semantically representative prototypes. For IPC=10 on ImageNette, this difference is visually evident: divisive-levelwise centroids align with class structure, whereas agglomerative ones drift toward noisy edges. \textit{\underline{Takeaway}:} Divisive clustering provides centrally anchored, density-aligned IPC centroids, while agglomerative clustering often produces edge-biased, noisier centroids.

\begin{figure}[!t]
    \centering
    \includegraphics[width=\linewidth, keepaspectratio]{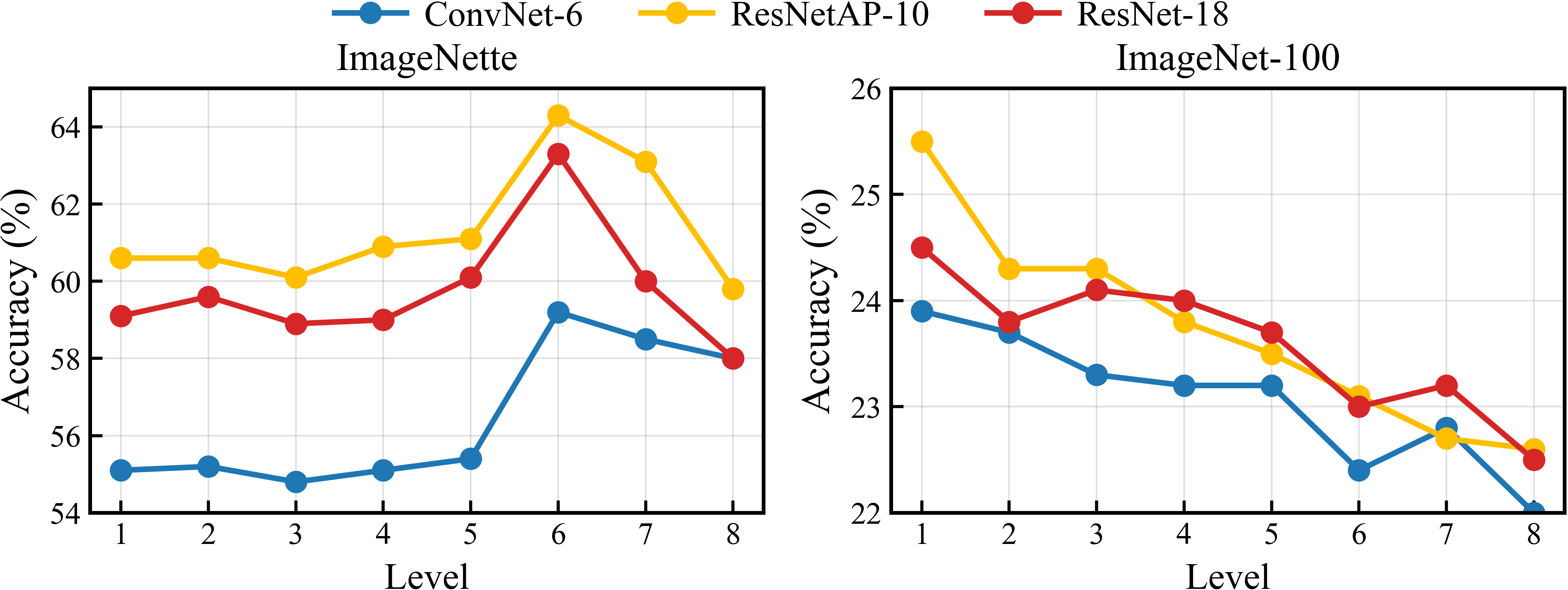}
    \caption{\footnotesize\textbf{Ablation experiments to find out the level of divisive-layerwise clustering.} It can be seen that for datasets with less class (ImageNette) the start level ($s_{\text{start}}$) is more whereas for datasets with more class (ImageNet-100) the level is less.} 
    \label{fig:ab_level}
\end{figure}

\noindent
\textbf{Level analysis for Divisive-levelwise clustering.} As shown in \cref{fig:ab_level}, ImageNette selects a higher $s_{\text{start}}$ compared to ImageNet-100, indicating that datasets with fewer classes benefit from starting IPC selection deeper in the hierarchy. With fewer classes, the VAE latent space is less congested, and class clusters occupy more separated, fine-grained regions, thus making leaf-level centroids reliable representatives. In contrast, ImageNet-100 exhibits substantial feature overlap due to its larger number of classes, causing leaf nodes to contain noisy, outlier-like samples that blur class boundaries. Starting closer to the root therefore avoids oversampling these unstable leaf clusters and yields more semantically coherent centroids. This validates the design of divisive-levelwise IPC selection, i.e., controlling $s_{\text{start}}$ naturally adapts the granularity of the centroid set to the intrinsic structure of each dataset’s feature manifold. \textit{\underline{Takeaway}:} The behavior of $s_{\text{start}}$ serves as a diagnostic for class-specific feature overlap. i) Low-overlap datasets (e.g., ImageNette): deeper, fine-level picks are effective. ii) High-overlap datasets (e.g., ImageNet-100): shallower levels preserve separability and avoid noise-dominated leaves.

\begin{table}[t]
\centering
\caption{\footnotesize\textbf{Effect of radius $r$ on downstream classification accuracy (\%).} Best result is highlighted with \textbf{bold}.}
%\vspace{0.3em}
\resizebox{\linewidth}{!}{%
\begin{tabular}{lccc|ccc}
\toprule
& \multicolumn{3}{c}{\textbf{ImageNette (IPC=10)}} 
& \multicolumn{3}{c}{\textbf{ImageNet-100 (IPC=10)}} \\
\cmidrule(lr){2-4}\cmidrule(lr){5-7}
\textbf{Classifier} & $r{=}0.05$ & $r{=}0.1$ & $r{=}0.2$ 
                    & $r{=}0.05$ & $r{=}0.1$ & $r{=}0.2$ \\
\midrule
ConvNet-6     & \textbf{60.5} & 59.7 & 59.5 & 23.7 & \textbf{24.5} & 24.0 \\
ResNetAP-10   & \textbf{63.7} & 63.2 & 63.0 & 26.5 & \textbf{27.4} & 27.2 \\
ResNet-18     & \textbf{62.3} & 62.0 & 61.2 & 23.1 & \textbf{23.8} & 23.4 \\
\bottomrule
\end{tabular}}
\label{tab:radius_ablation}
\end{table}

\noindent
\textbf{Effect of Neighborhood Radius $r$ in Local Manifold Estimation.} In our method, each IPC centroid $c_s$ defines a local latent neighborhood $\mathcal{N}_s = \{ z \in Z : \|z - c_s\|_2 \le r \}$, where $r$ is a radius in the \emph{VAE latent space}.  
Intuitively, a smaller $r$ yields a tightly localized neighborhood capturing fine-grained geometric structure, while a larger $r$ aggregates a broader region of the latent feature cloud, mixing together points with potentially different local curvature or class-specific variations.  
Since the tangent space and normal projector are computed from the covariance of $\mathcal{N}_s$, the choice of $r$ directly affects the quality of the estimated local diffusion manifold $\mathcal{M}_t^{(s)}$. For ImageNette (10-class subset), we observe that a small radius $r=0.05$ performs best, while performance degrades for $r=0.1$ and $r=0.2$. Conversely, for ImageNet-100 (100 classes), the trend reverses: $r=0.05$ tends to fail, and moderate radii $r=0.1$ or $r=0.2$ yield better performance. This pattern is expected because $r$ interacts with \emph{dataset granularity}, \emph{class overlap}, and \emph{latent density}. In ImageNette, classes occupy well-separated, compact regions of the VAE space, so a small radius correctly isolates coherent local geometry. In ImageNet-100, however, the latent space becomes more crowded (classes have heavier overlap and higher intrinsic variation) so $r=0.05$ often yields too few neighbors to estimate a stable covariance, causing unreliable tangent spaces. Moderate radii provide enough samples to obtain statistically stable, noise-robust local geometry. \textit{\underline{Takeaway}:}
$r$ must match the density and geometry of the feature space. Small $r$ excels when class clusters are compact and well-separated (e.g., ImageNette). Larger $r$ becomes necessary in denser, higher-class settings (e.g., ImageNet-100) to obtain stable covariance and reliable tangent estimation. This reinforces that effective manifold guidance depends on capturing \emph{locally coherent} structure, i.e., not too narrow (noisy), not too broad (geometrically inconsistent).

\begin{table}[t]
\centering
\small
\caption{\footnotesize\textbf{Performance comparison using DDIM.} Using ImageNette IPC=10 setting for comparison.}
\begin{tabular}{l|ccc}
\hline
Method & ResNet-18 & ResNet-10 & ConvNet-6 \\
\hline
MGD \cite{chan2025mgd3}    & 64.6 & 62.7 & 60.8 \\
ManifoldGD & 66.5 (\textcolor{green}{+1.9}) & 63.6 (\textcolor{green}{+0.9}) & 62.1 (\textcolor{green}{+1.3}) \\
\hline
\end{tabular}
\label{tab:ddim}
\end{table}

\noindent
\textbf{Effect of Tangent space dimension $d$ in Local Manifold Estimation.} Tab. \cref{tab:ablation_d} analyzes the effect of tangent subspace dimension d. We observe a consistent peak at d=3. $d<3$ under-represents the local manifold, restricting valid tangent directions. $d>3$ introduces noisy/off-manifold directions. \textit{\underline{Takeaway}:} Performance is stable across a reasonable range of d, indicating low sensitivity to this hyperparameter. This is expected since manifold correction is local and only requires principal variation directions (not full geometric reconstruction).

\begin{table}[t]
\centering
\caption{Effect of $d$ on ConvNet-6 performance (IPC=10), transposed.}
\begin{tabular}{c|ccccc}
\toprule
$d$ & 1 & 2 & 3 & 4 & 5 \\
\midrule
Nette & 58.6 & 58.5 & \textbf{59.5} & 58.2 & 57.9 \\
ImageNet-100 & 24.0 & 24.1 & \textbf{24.5} & 24.4 & 23.8 \\
\bottomrule
\end{tabular}
\label{tab:ablation_d}
\end{table}

\noindent
\textbf{Comparison using different schedulers.} From \cref{tab:ddim} we can see that ManifoldGD outperforms MGD \cite{chan2025mgd3} while using a DDIM noise scheduler as well. It was already seen in \cref{tab:comparison,tab:imagewoof} that ManifoldGD achieves better performance with DDPM scheduler as well. \textit{\underline{Takeaway}:} Superior performance of ManifoldGD over training-free methods is scheduler agnostic.

\begin{figure}[!t]
    \centering
    \includegraphics[width=0.8\linewidth, keepaspectratio]{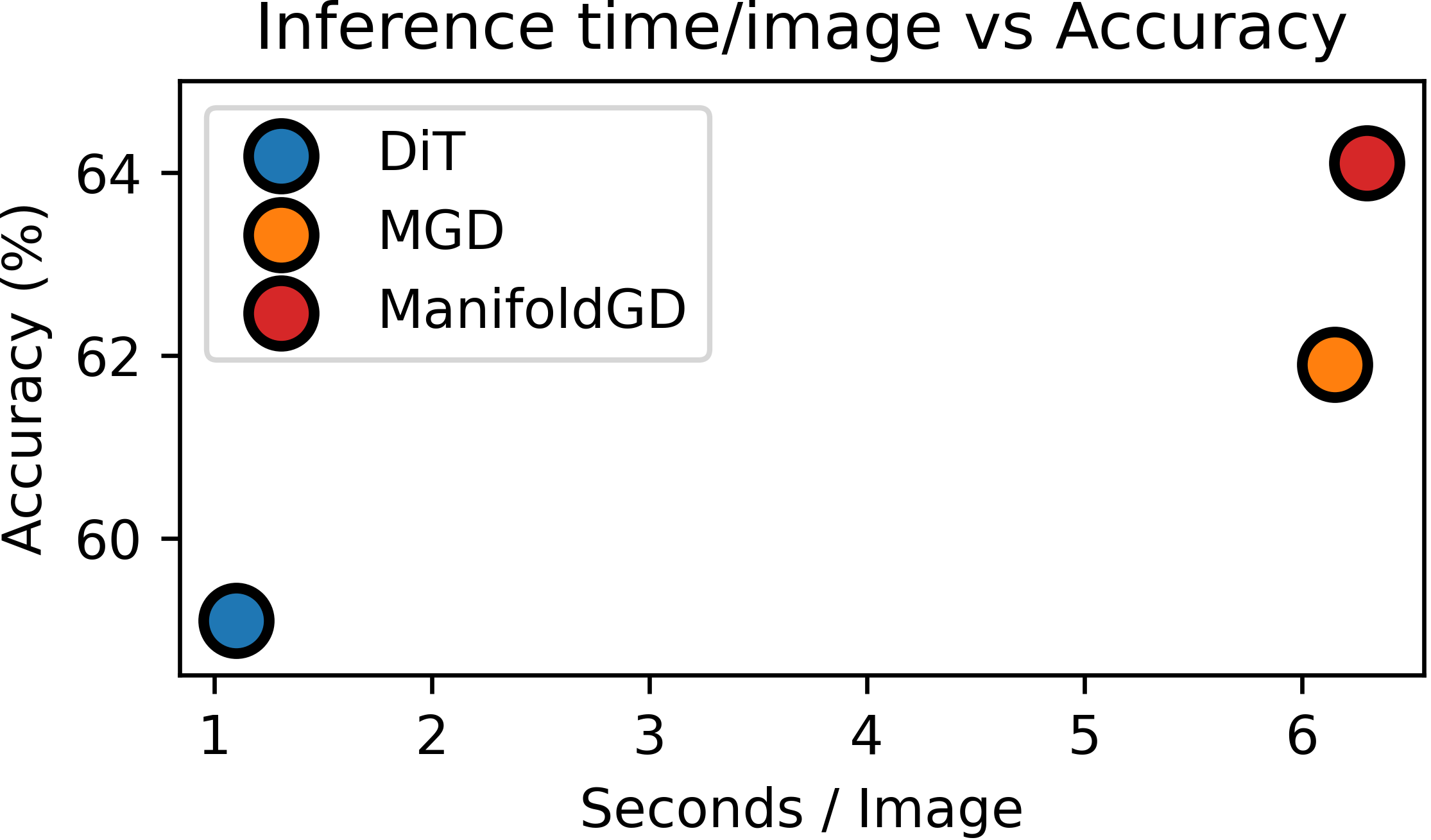}
    \vspace{-1.0em}\caption{\footnotesize\textbf{Inference time vs Accuracy of MGD \cite{chan2025mgd3}, DiT \cite{peebles2023dit} and ManifoldGD for ImageNette IPC 10.} We see that ManifoldGD has a higher inference compared to MGD \cite{chan2025mgd3} and DiT \cite{peebles2023dit} while also producing superior performance.} 
    \label{fig:time_complexity}
    \vspace{-1.5em}
\end{figure}

\noindent
\textbf{Time complexity/comparison.} We measure inference time for training-free methods when generating one ImageNette sample (IPC=10) on an NVIDIA A6000 (Fig. \cref{fig:time_complexity}) ManifoldGD adds modest overhead compared to pure inference-time guidance, yet remains fully training-free, which would otherwise have incurred substantially higher compute and memory costs. In practice, ManifoldGD trades a small increase in sampling time (-$\Delta0.15s$ vs. MGD [34], -$\Delta5.2s$ vs. DiT [28]) for improved data quality and downstream performance (+$\Delta2.2$ vs. MGD [34], +$\Delta5.0$ vs. DiT [28]).

\begin{figure}[!t]
    \centering
    \includegraphics[width=\linewidth, keepaspectratio]{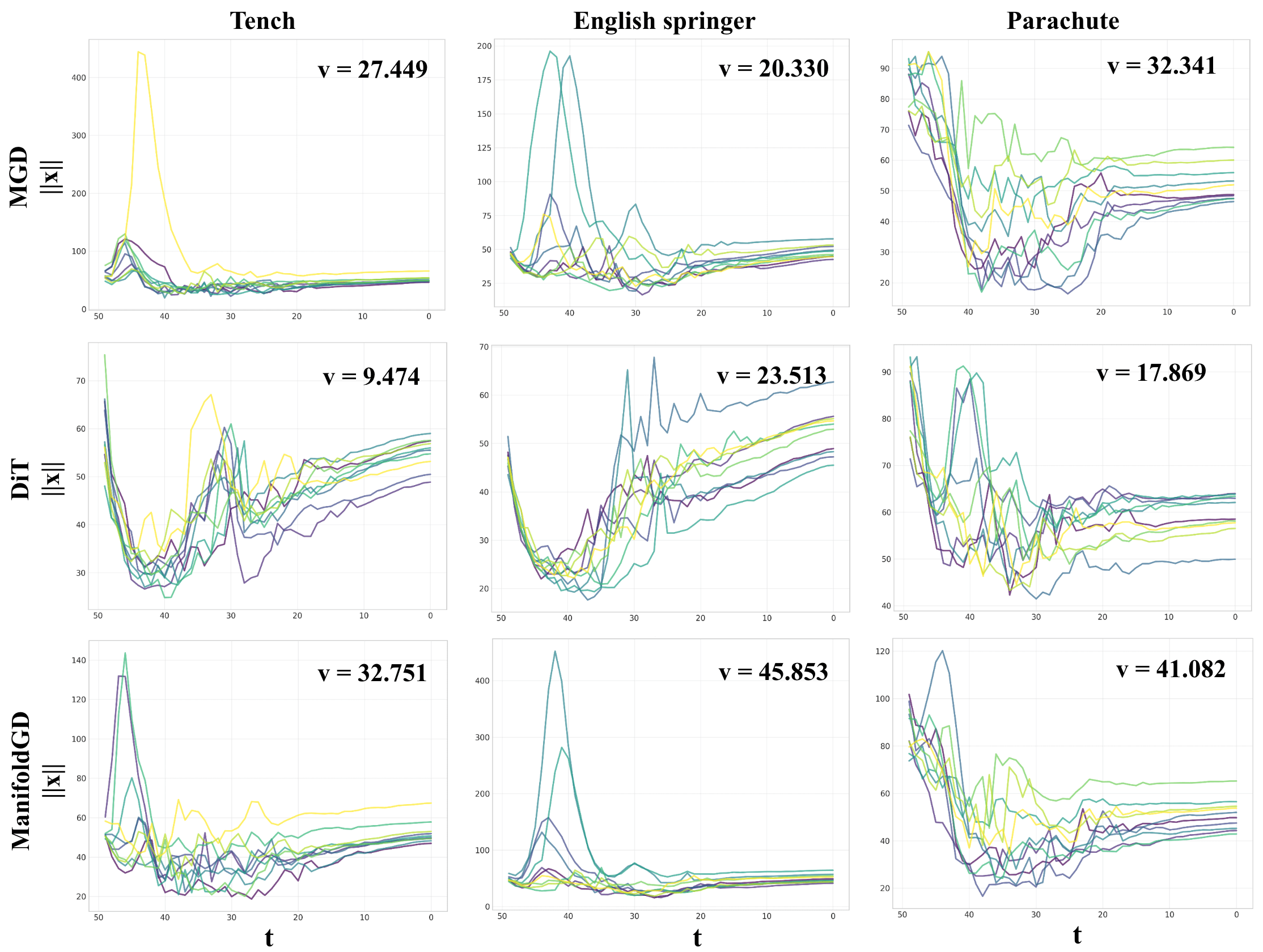}
    \vspace{-1.0em}\caption{\footnotesize\textbf{Trajectory comparison of MGD \cite{chan2025mgd3}, DiT \cite{peebles2023dit} and ManifoldGD for ImageNette IPC 10.} We see that ManifoldGD has a higher final trajectory variance of the norm of the latent vector $||x||$ compared to MGD \cite{chan2025mgd3} and DiT \cite{peebles2023dit}.} 
    \label{fig:trajectories}
    \vspace{-1.5em}
\end{figure}

\noindent
\textbf{Analysis of denoising trajectories.} From \cref{tab:ddim}. We visualize the per-step denoising trajectories by tracking the Euclidean norm of the latent prediction $||\hat{x}||$ (the VAE-latent decoded estimate returned by the diffusion step) for multiple IPC centroid-conditioned samples (10 trajectories for IPC=10) and computing the variance of those norms across samples at each timestep. Concretely, for each conditioned sample we record $||\hat{x}(t)||$ at every denoising step, then plot the sample-wise trajectories (left) and the timestep-wise variance of those norms (right). The plotted final variance (v) refers to the variance of $||\hat{x}(50)||$ at the end of the reverse process (t=50). \cref{fig:trajectories} shows these trajectories for DiT, MGD, and ManifoldGD on ImageNette (IPC=10). ManifoldGD consistently produces a larger final trajectory variance than both MGD \cite{chan2025mgd3} and DiT \cite{peebles2023dit}. This is an expression of meaningful intra-mode variability preserved by our manifold correction. Intuitively, MGD’s \cite{chan2025mgd3} plain Euclidean mode-pull strongly concentrates samples toward the centroid direction and therefore reduces final variability (mode-collapse within a prototype). DiT \cite{peebles2023dit} being unguided often yields lower variance as well because it samples broadly from the prior but lacks targeted, class-anchored diversity. By contrast, ManifoldGD combines semantic attraction (mode guidance) with an explicit cancellation of the off-manifold normal component, i.e., we remove the normal projection and keep tangent-aligned variations. This lets samples remain close to class modes while preserving the valid degrees of freedom along the manifold (higher tangent-aligned spread), which increases the measured final variance. \textit{\underline{Takeaway}:} ManifoldGD’s larger final trajectory variance reflects useful, manifold-consistent intra-class diversity rather than noise. By eliminating the off-manifold normal component while retaining tangent-aligned variations, ManifoldGD avoids the two failure modes of plain mode-pulling (i) collapse onto a tight prototype and (ii) drift off the data manifold, so samples remain class-relevant yet diverse. Because FID and downstream accuracy also improve, the increased variance serves as a compact diagnostic of the desired balance between representativeness and geometric fidelity.

\noindent
\textbf{Class-wise analysis.}
\cref{tab:imagewoof_classwise_rotated} presents a detailed class-wise comparison between MGD and ManifoldGD on ImageWoof for IPC = 10 and IPC = 50 across three architectures. ImageWoof is intentionally challenging for dataset distillation because all classes are dog breeds with highly overlapping visual structure (textures, fur patterns, pose similarity), making the latent space significantly more homogeneous than ImageNette or ImageNet-100. In such settings, mode-collapse and insufficient intra-class variation are more likely, and class separation depends heavily on preserving subtle geometric cues. This makes ImageWoof an especially sensitive benchmark for evaluating whether a distillation method can retain fine-grained, within-class diversity. Across IPC = 10, ManifoldGD improves over MGD on a majority of classes and architectures. For example, on ResNet-AP, ManifoldGD achieves notable gains on Beagle (+3.1), Shih-Tzu (+10.5), Ridgeback (+7.2), and Golden Retriever (+11.2), showing that manifold-aligned variation helps recover class-specific fine-grained details that Euclidean mode-pulling suppresses. Similar trends appear for ConvNet (Beagle +12.9, Samoyed +4.8) and ResNet-18 (Samoyed +4.5, Beagle +6.7, Ridgeback +9.8). These improvements emerge specifically in classes with subtle visual boundaries, where preserving tangent directions of the manifold is more critical than strict centroid attraction. At IPC = 50, both methods improve (as expected with increased supervision) but ManifoldGD continues to outperform MGD on most architectures. In particular, for ResNet-18, large gains appear for Beagle (+13.9), Shih-Tzu (+3.2), Ridgeback (+5.9), and Golden Retriever (+4.1). ConvNet and ResNet-AP show similar improvements, especially on texture-heavy breeds such as Shih-Tzu, Ridgeback, and Golden Retriever. These gains indicate that even with a larger IPC budget, enforcing manifold-consistent guidance preserves meaningful degrees of freedom within each breed’s local geometry, preventing over-regularization and enabling better fine-grained discrimination. \textit{\underline{Takeaway}:} ManifoldGD is especially effective on fine-grained, homogeneous datasets like ImageWoof, where class distinctions rely on subtle manifold-level variations rather than coarse semantic separation. The consistent improvements across architectures and IPC budgets demonstrate that combining semantic attraction with manifold-consistent tangent preservation yields more discriminative and diverse synthetic data, ultimately improving downstream classification accuracy.

\section{Qualitative Analysis}
\label{sec:additional_qualitative}

\begin{figure*}[!t]
    \centering
    \includegraphics[width=\linewidth, keepaspectratio]{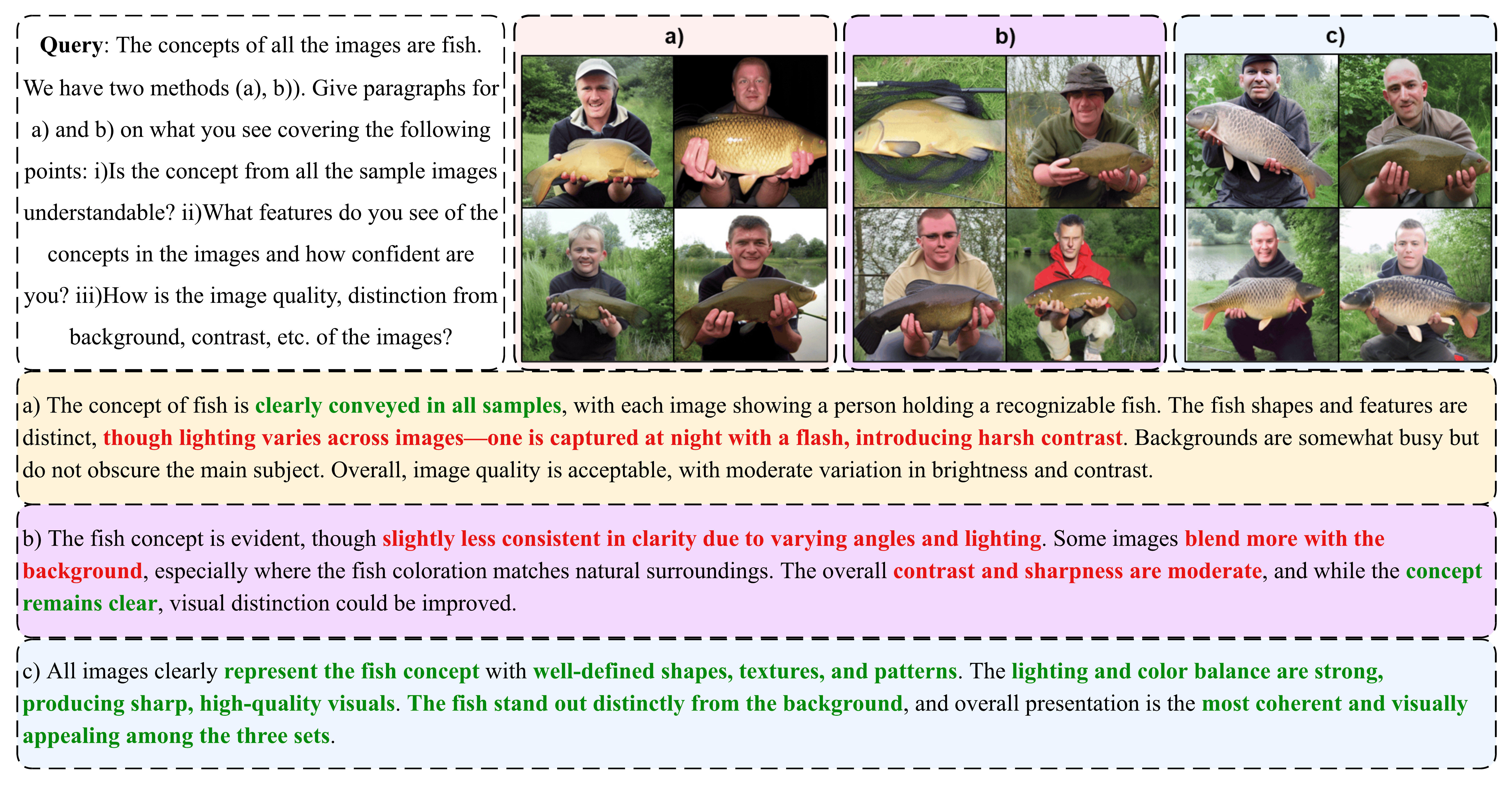}
    \vspace{-1.0em}\caption{\footnotesize\textbf{Qualitative comparison of MGD \cite{chan2025mgd3}, DiT \cite{peebles2023dit} and ManifoldGD.} We provide the Query to instruct GPT 4.o to evaluate the images (a) $\to$ DiT, b) $\to$ MGD, c) $\to$ ManifoldGD). We see from the answers of GPT 4.o that ManifoldGD achieves better images that discriminate the concept (fish) from other information (background) better. Green highlighted text indicates positive attributes whereas Red highlighted text indicates negative attributes.} 
    \label{fig:compare_data_1}
    \vspace{-1.0em}
\end{figure*}

\begin{figure*}[!t]
    \centering
    \includegraphics[width=\linewidth, keepaspectratio]{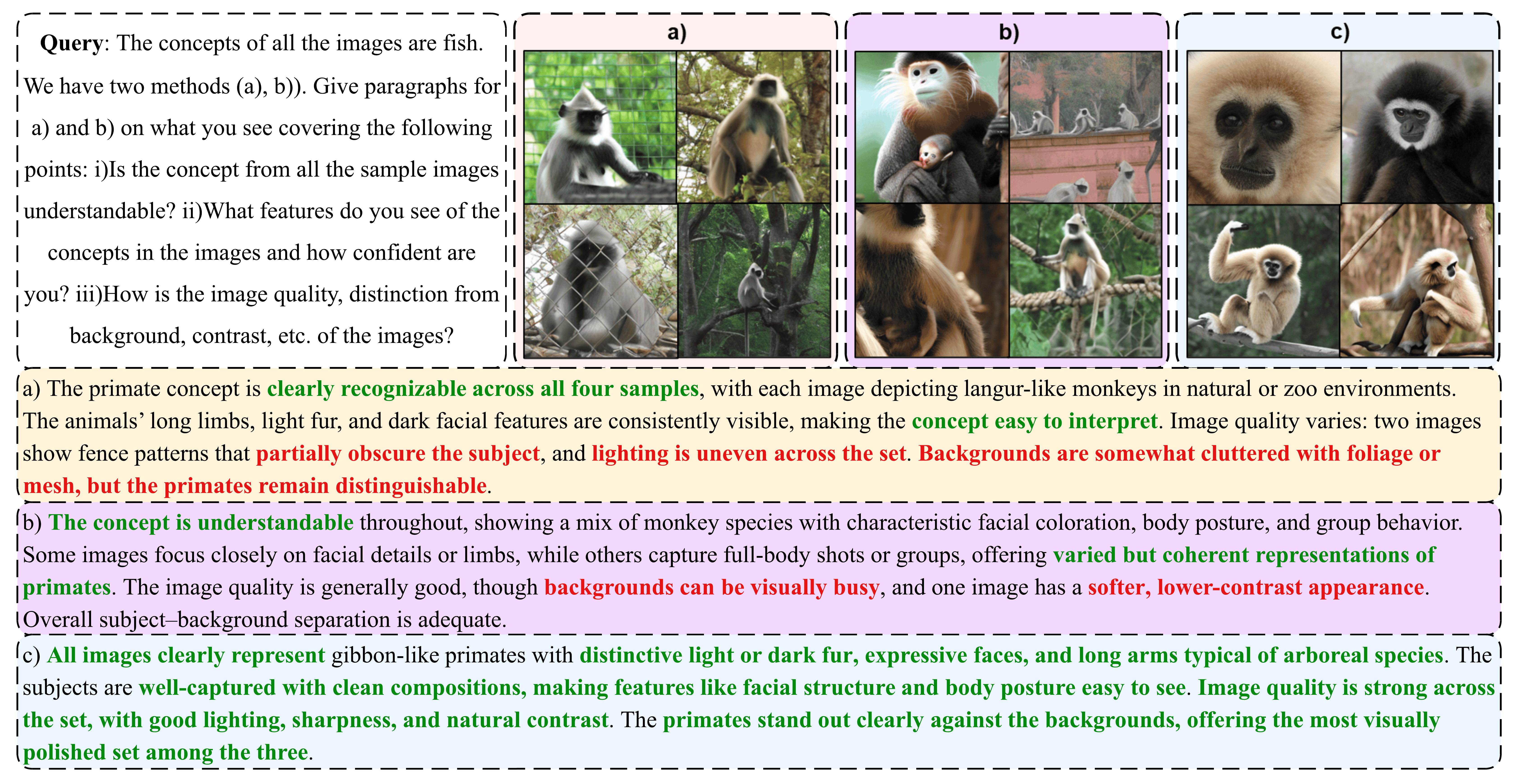}
    \vspace{-1.0em}\caption{\footnotesize\textbf{Qualitative comparison of MGD \cite{chan2025mgd3}, DiT \cite{peebles2023dit} and ManifoldGD.} We provide the Query to instruct GPT 4.o to evaluate the images (a) $\to$ DiT, b) $\to$ MGD, c) $\to$ ManifoldGD). We see from the answers of GPT 4.o that ManifoldGD achieves better images that discriminate the concept (monkey) from other information (background) better. Green highlighted text indicates positive attributes whereas Red highlighted text indicates negative attributes.} 
    \label{fig:compare_data_2}
    \vspace{-1.5em}
\end{figure*}

\cref{fig:compare_data_1,fig:compare_data_2} shows a qualitative comparison between MGD~\cite{chan2025mgd3}, DiT \cite{peebles2023dit} and ManifoldGD. GPT-4.0 was used as an unbiased evaluator, prompted to assess conceptual clarity, visual sharpness, and background separation without revealing method identities. The responses indicate that ManifoldGD generates sharper, more coherent images where the concept “fish” and “monkey” are clearly distinguished from the background, exhibiting better texture and lighting consistency. In contrast, MGD \cite{chan2025mgd3} samples display moderate sharpness and weaker subject–background separation. This supports that $g_t^{\mathrm{man}}$ promotes denoising along intrinsic manifold directions, yielding semantically faithful yet geometrically consistent generations. Furthermore, ManifoldGD achieves variations (See \cref{fig:compare_data_2}) without compromising image quality which is lacking for DiT \cite{peebles2023dit} and MGD \cite{chan2025mgd3}.

\begin{figure*}[!t]
    \centering
    \includegraphics[width=\linewidth, keepaspectratio]{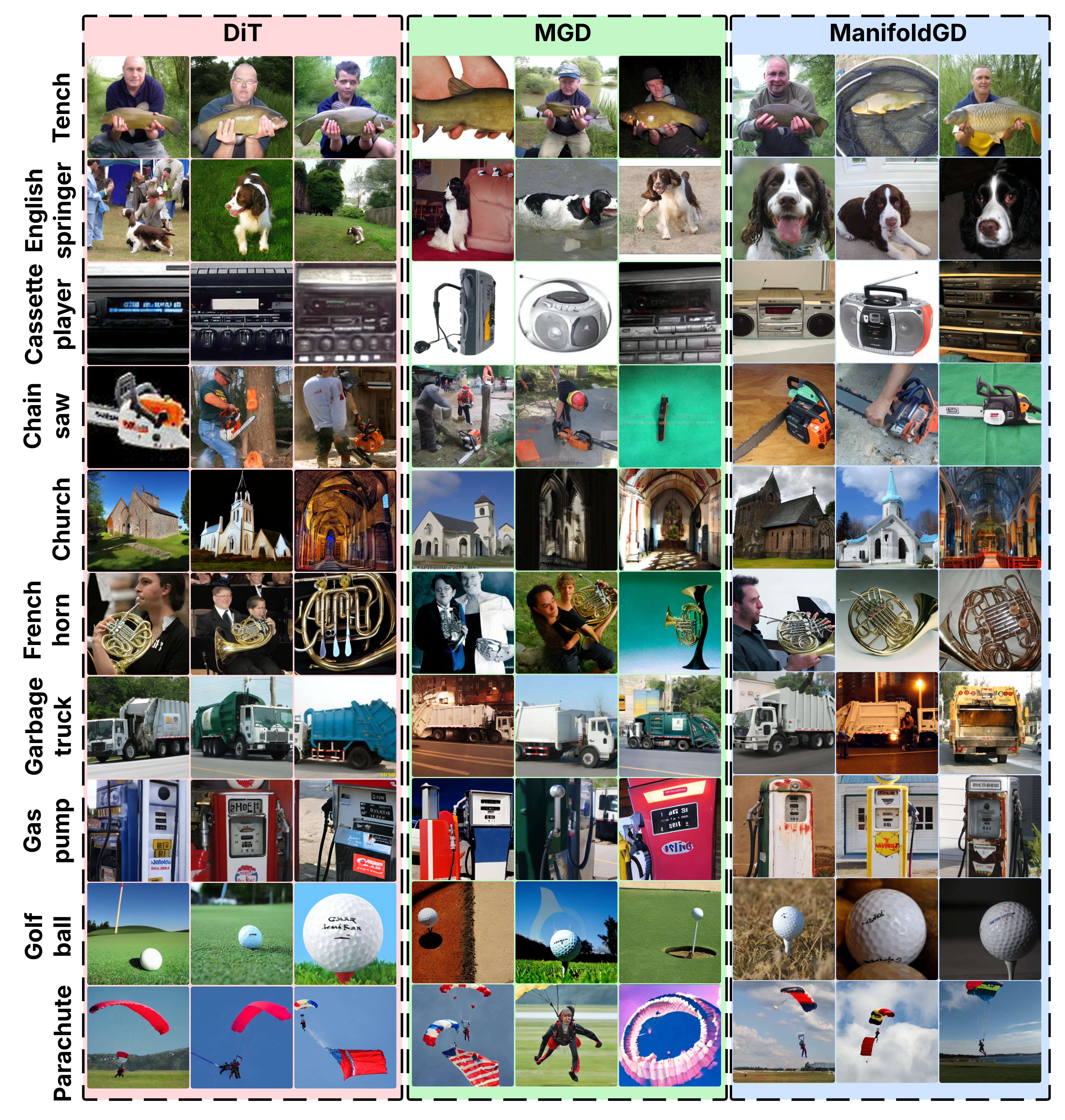}
    \vspace{-1.0em}\caption{\footnotesize\textbf{Qualitative comparison of MGD \cite{chan2025mgd3}, DiT \cite{peebles2023dit} and ManifoldGD for ImageNette.} 
    ManifoldGD produces sharper, more semantically aligned and structurally coherent samples compared to MGD, while also avoiding the occasional blurring or texture flattening observed in DiT. 
    Differences are especially prominent in edges, fine textures (fur, feathers), and object–background boundaries.}
    \label{fig:qualitative_nette}
    \vspace{-1.5em}
\end{figure*}

\begin{figure*}[!t]
    \centering
    \includegraphics[width=\linewidth, keepaspectratio]{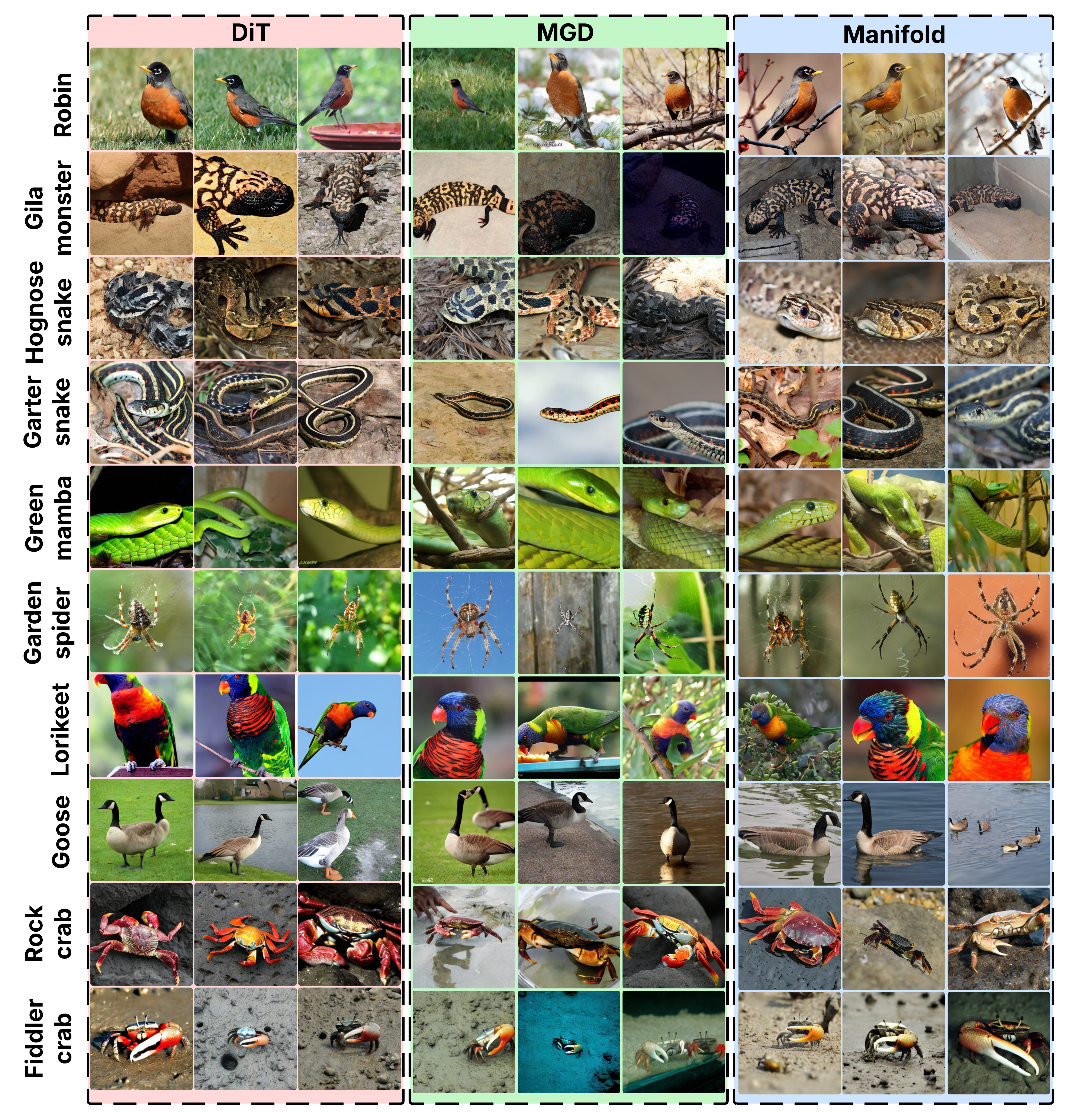}
    \vspace{-1.0em}\caption{\footnotesize\textbf{Qualitative comparison of MGD \cite{chan2025mgd3}, DiT \cite{peebles2023dit} and ManifoldGD for ImageNet-100.} 
    All three training-free methods generate visually plausible samples for this larger and more diverse dataset, though ManifoldGD still achieves cleaner geometry, fewer artifacts, and improved object–background separation. 
    Differences are more subtle than in ImageNette due to the higher visual entropy and class variability of ImageNet-100.} 
    \label{fig:qualitative_imagenet-100}
    \vspace{-1.5em}
\end{figure*}

\cref{fig:qualitative_nette,fig:qualitative_imagenet-100} compare DiT \cite{peebles2023dit}, MGD \cite{chan2025mgd3}, and ManifoldGD samples on ImageNette and ImageNet-100.  
For ImageNette, the visual gap between methods is more pronounced: MGD often exhibits over-smoothed textures or centroid-induced collapse, while DiT occasionally drifts off-class or produces weaker semantic detail. ManifoldGD, by contrast, consistently produces sharper boundaries, richer textures, and more stable object geometry reflecting the benefit of tangent-aligned guidance and the removal of off-manifold drift. For ImageNet-100, all methods generate reasonably strong images because the underlying data distribution is more visually diverse and less constrained, making minor errors less perceptually salient. Nevertheless, ManifoldGD still improves FID (see \cref{fig:fid_rep_div_comparison}) and yields slightly cleaner, more coherent samples, even though the perceptual gap appears narrower. Overall, the qualitative results mirror our quantitative trends: on simpler, low-entropy datasets (ImageNette), manifold-corrected guidance yields visibly stronger structure and detail, while on high-entropy datasets (ImageNet-100), improvements remain measurable but more subtle.

\section{Results on ImageNet-1k}
\label{sec:additional_benchmark}

We additionally benchmark DiT \cite{peebles2023dit}, MGD \cite{chan2025mgd3}, and ManifoldGD on the full ImageNet-1k dataset under the most challenging hard-label evaluation protocol for IPC = 1 and 50. As shown in \cref{tab:net1k_suppl_clean}, ManifoldGD consistently outperforms both baselines across both IPC regimes. These gains reflect the effectiveness of our manifold-aligned guidance in preserving class-consistent structure even when only a single distilled exemplar per class is available.

Furthermore, ImageNet-1k represents a significantly more challenging testbed compared to ImageNette or ImageNet-100. The dataset contains 1000 classes, many of which are fine-grained, visually similar, and highly imbalanced in the VAE latent space. This dramatically increases mode density and inter-class overlap, making it difficult for training-free methods to avoid collapsing onto overly generic prototypes. Methods relying solely on Euclidean mode attraction (e.g., MGD \cite{chan2025mgd3}) tend to over-concentrate samples around coarse class directions, while unguided diffusion sampling (DiT \cite{peebles2023dit}) lacks the semantic anchoring required to generate discriminative exemplars. In contrast, ManifoldGD’s tangent-aligned guidance preserves intra-class variability while suppressing off-manifold drift, leading to more faithful and well-separated prototypes across a large number of classes.

Overall, these results demonstrate that ManifoldGD is not only effective in small-scale settings but also scales robustly to the full ImageNet-1k challenge, maintaining strong sample fidelity and discriminative power even at extremely low IPC budgets.

\begin{table}[t]
\centering
\small
\caption{\textbf{Comparison on ImageNet-1k using ConvNet-6.} Best results are in \textbf{bold}, second-best are \underline{underlined}. * are results of methods achieved after reimplementation}
\vspace{0.4em}
\label{tab:net1k_suppl_clean}

\resizebox{0.8\linewidth}{!}{
\begin{tabular}{c|c|c|c}
\toprule
\textbf{IPC} &
\textbf{DiT*\cite{peebles2023dit}} &
\textbf{MGD*\cite{chan2025mgd3}} &
\textbf{ManifoldGD} \\
\midrule

1 
& $2.6_{\tiny\pm0.3}$ 
& \underline{$2.8_{\tiny\pm0.9}$} 
& \textbf{3.1$_{\tiny\pm0.9}$} (\textcolor{green}{+0.3\%}) \\

50 
& $18.5_{\tiny\pm1.3}$ 
& \underline{$20.3_{\tiny\pm1.1}$} 
& \textbf{21.4$_{\tiny\pm1.5}$} (\textcolor{green}{+1.1\%}) \\
\bottomrule
\end{tabular}
}
\end{table}

\end{document}